%% file: acl_latex.tex
\pdfoutput=1
\documentclass[11pt]{article}

\usepackage[preprint]{acl}

\usepackage{times}
\usepackage{latexsym}
\usepackage{xspace}
\usepackage{amsmath}
\usepackage[most]{tcolorbox}
\usepackage{enumitem}
\usepackage[T1]{fontenc}
\usepackage[utf8]{inputenc}

\usepackage{microtype}

\usepackage{inconsolata}

\usepackage{graphicx}
\usepackage{booktabs}
\usepackage{multirow}
\usepackage{soul}
\usepackage{xcolor}

\newcommand{\method}{{\textsc{Magi}}\xspace}
\colorlet{soulyellow}{yellow!50}
\colorlet{soulorange}{orange!30}
\colorlet{soulblue}{cyan!20}
\colorlet{soulgreen}{green!20}
\title{\method: \underline{M}ulti-\underline{A}gent \underline{G}uided \underline{I}nterview for Psychiatric Assessment}

\author{
 \textbf{Guanqun Bi\textsuperscript{1}},
 \textbf{Zhuang Chen\textsuperscript{1,2}},
 \textbf{Zhoufu Liu\textsuperscript{3}},
 \textbf{Hongkai Wang\textsuperscript{4}},
\\
 \textbf{Xiyao Xiao\textsuperscript{5}},
 \textbf{Yuqiang Xie\textsuperscript{6}},
 \textbf{Wen Zhang\textsuperscript{7}},
 \textbf{Yongkang Huang\textsuperscript{5}},
\\
 \textbf{Yuxuan Chen\textsuperscript{1}},
 \textbf{Libiao Peng\textsuperscript{5}},
 \textbf{Yi Feng\textsuperscript{8}},
 \textbf{Minlie Huang\textsuperscript{1}}
\\
\\
 \textsuperscript{1}CoAI Group, DCST, IAI, BNRIST, Tsinghua University,
 \textsuperscript{2}Central South University,\\
 \textsuperscript{3}Beijing Institute of Technology,
 \textsuperscript{4}Sichuan University,
 \textsuperscript{5}Lingxin AI, \\
  \textsuperscript{6}Independent Researcher,
\textsuperscript{7}University of International Relations,\\
\textsuperscript{8}Central University of Finance and Economics
\\
{biguanqun@mail.tsinghua.edu.cn \quad aihuang@tsinghua.edu.cn}}

\begin{document}
\maketitle
\begin{abstract}
Automating structured psychiatric interviews could revolutionize mental healthcare accessibility, yet existing large language models (LLMs) approaches fail to align with psychiatric diagnostic protocols. 
We present \method, the first framework that transforms the gold-standard Mini International Neuropsychiatric Interview (MINI) into automatic computational workflows through coordinated multi-agent collaboration. 
\method dynamically navigates clinical logic via four specialized agents: 1) an interview tree guided navigation agent adhering to the MINI's branching structure, 2) an adaptive question agent blending diagnostic probing, explaining, and empathy, 3) a judgment agent validating whether the response from participants meet the node, and 4) a diagnosis agent generating Psychometric Chain-of-Thought (PsyCoT) traces that explicitly map symptoms to clinical criteria. 
Experimental results on 1,002 real-world participants covering depression, generalized anxiety, social anxiety, and suicide show that \method advances LLM-assisted mental health assessment by combining clinical rigor, conversational adaptability, and explainable reasoning.

\end{abstract}

\section{Introduction}

Mental health disorders pose a significant global challenge, and accurate and timely diagnosis is crucial for effective intervention \cite{apa2013dsm}.  
Clinical interviews, particularly structured instruments, are gold standards for psychiatric assessment \cite{crisp2014global}.  
The Mini International Neuropsychiatric Interview (MINI) \cite{meyer2001psychological} is a widely adopted structured interview protocol, providing a systematic approach to assess major psychiatric disorders.
However, their manual administration is time-consuming and resource-intensive, creating a bottleneck in mental healthcare access and limiting the widespread use in real-world settings.

\begin{figure}[tbp]
	\centering
	\includegraphics[width=0.49\textwidth]{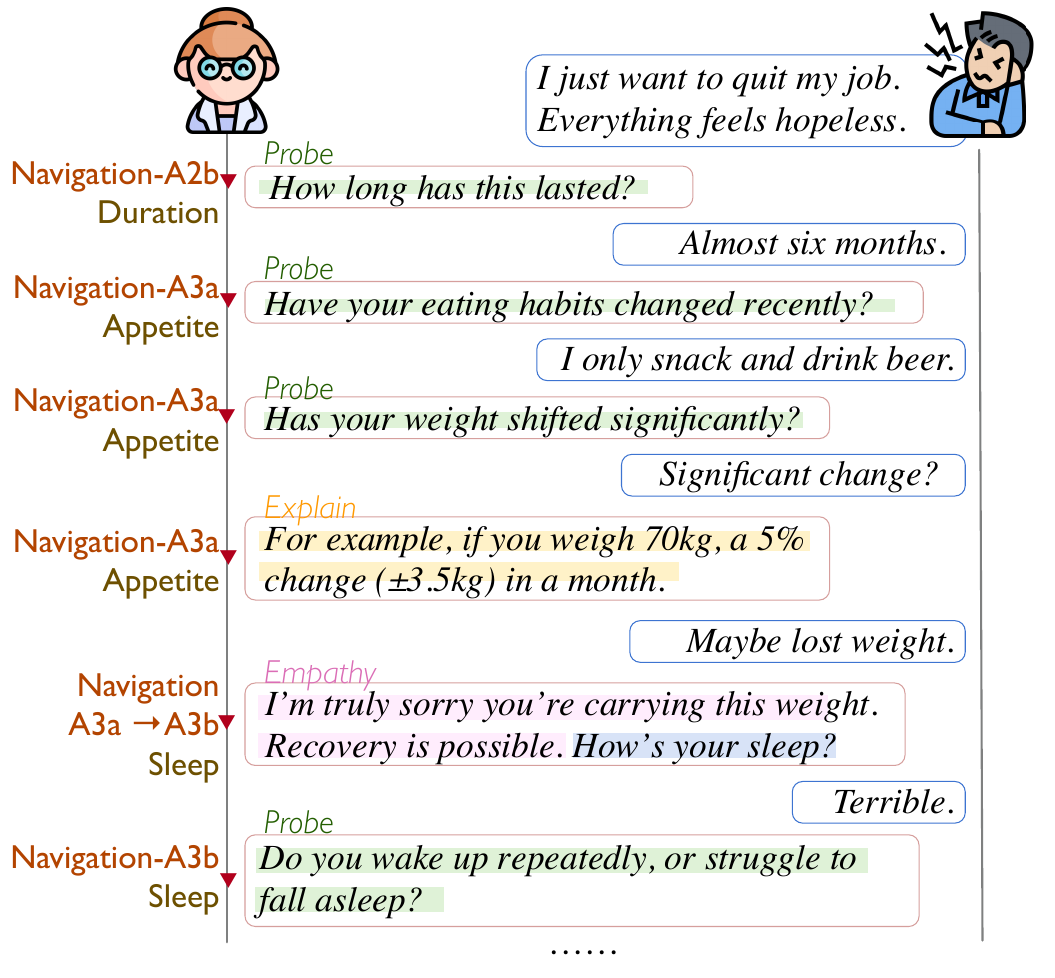}
	\caption{Example dialogue flow from \method. Our multi-agent framework \method guides participants through structured psychiatric interviews following MINI protocol.
    }
	\label{fig-example}
\vspace{-0.5em}
\end{figure}

The advent of large language models (LLMs) \cite{openai2023gpt4, touvron2023llama, deepseek2024deepseek,deepseekai2025deepseekr1incentivizingreasoningcapability} offers promising avenues to automate and enhance mental health support processes, potentially democratizing access to care \cite{he2023towards, zhao2023survey}. 
While LLM-based mental health tools demonstrate conversational fluency \cite{tu2024towards, sartori2023language}, three fundamental mismatches with clinical practice impede their adoption. 
\textbf{(1) Diagnostic framework deviation}: Current open-ended LLM dialogues frequently diverge from established clinical standards like DSM-5, particularly struggling with nuanced differential diagnoses \cite{apa2013dsm}.
\textbf{(2) Rigid interaction patterns}: Existing systems employ inflexible questioning templates that prove inadequate when participants exhibit emotional distress or require clarifications, contradicting evidence-based interviewing principles \cite{milintsevich2023towards, zhang2024multilevel}. 
\textbf{(3) Limited diagnosis transparency}: Most systems operate as ``black boxes'', providing little insight into their diagnostic reasoning process. This opacity significantly undermines trust and hinders clinical adoption, especially in high-stakes healthcare settings \cite{sadeghi2023exploring}.

To bridge these gaps, we present \method (\underline{M}ulti-\underline{A}gent \underline{G}uided \underline{I}nterview), a clinical logic-driven framework that operationalizes the MINI interview through four coordinated agents.
%\cite{chen2023autoagents}. 
Consider a depressive episode screening shown in Figure~\ref{fig-example}: when a participant reports feeling hopeless, the navigation agent activates the diagnostic nodes systematically from duration (A2b) to appetite (A3a) and sleep (A3b), dynamically managing interview progression. Guided by this structure, the question agent adaptively switches between three strategies: structure-driven probing (``How long has this lasted?''), explaining for ambiguous concepts (clarifying significant weight change), and empathetic support when detecting distress \cite{sharma2023cognitive}. The judgment agent continuously validates responses against MINI criteria, iterating until all necessary evidence is collected. Finally, the diagnosis agent synthesizes the accumulated evidence into formal diagnostic conclusions through Psychometric Chain-of-Thought (PsyCoT), a clinical reasoning paradigm that explicitly connects symptom observations to diagnostic criteria via intermediate psychiatric constructs.

Through a school-based study, we collect 
%and expertly annotated 
1,002 clinical interviews covering depression, generalized anxiety, social anxiety, and suicide risk. 
With dual-expert annotation from licensed psychologists, our experiments demonstrate \method's clinical validity, achieving high agreement with expert diagnoses while providing transparent diagnostic traces absent in prior LLM approaches \cite{galatzer2023capability, chen2023empowering}.

Our contributions are threefold:

(1) \textbf{Clinical-Grounded Architecture}: We present a novel multi-agent framework \method that effectively embodies the procedural logic of the MINI within a dynamic semi-structured dialogue system, advancing the state-of-the-art in automated clinical interviews.

(2) \textbf{Explainable Diagnostic Reasoning}: We propose the Psychometric Chain-of-Thought (PsyCoT) reasoning paradigm, which enhances the transparency and clinical interpretability of AI-driven diagnostic processes by explicitly mirroring psychological measurement principles.

(3) \textbf{Large-Scale Clinical Validation}: Experiments through a school-based study with 1,002 participants and dual-expert clinical annotation demonstrates the framework's superior diagnostic performance against baseline methods.

\section{Related Work}
\subsection{Automatic Mental Health Assessment}

Traditional psychiatric assessment methods, including standardized questionnaires and expert interviews \cite{meyer2001psychological}, face engagement and accessibility challenges \cite{merry2012effectiveness}. To address these limitations, researchers have explored computerized adaptive testing \cite{meijer1999computerized} and game-based electronic assessments \cite{jones1984video,kim2016applying} to facilitate intrinsic motivation while maintaining assessment validity. Recent natural language processing advances have enabled sophisticated analysis of mental health indicators \cite{sharma2020computational,sharma2023cognitive}, while LLMs have demonstrated potential in cognitive distortion detection \cite{chen2023empowering}, psychologist role-playing \cite{tu2024towards}, and clinical interview analysis \cite{galatzer2023capability}.

LLM-based multi-agent systems have proven effective in complex tasks requiring coordination \cite{park2023generative}. However, existing LLM-based psychological assessment approaches often struggle with generalizability across psychological constructs \cite{chen2023empowering} and maintaining consistent engagement \cite{tu2024towards}. Our work introduces a novel multi-agent framework that combines gamification's engagement benefits with LLMs' flexibility and capabilities, addressing both generalizability across psychological constructs and participant engagement through interactive narrative generation.

\begin{figure*}[t] 
    \centering
    \includegraphics[width=\linewidth]{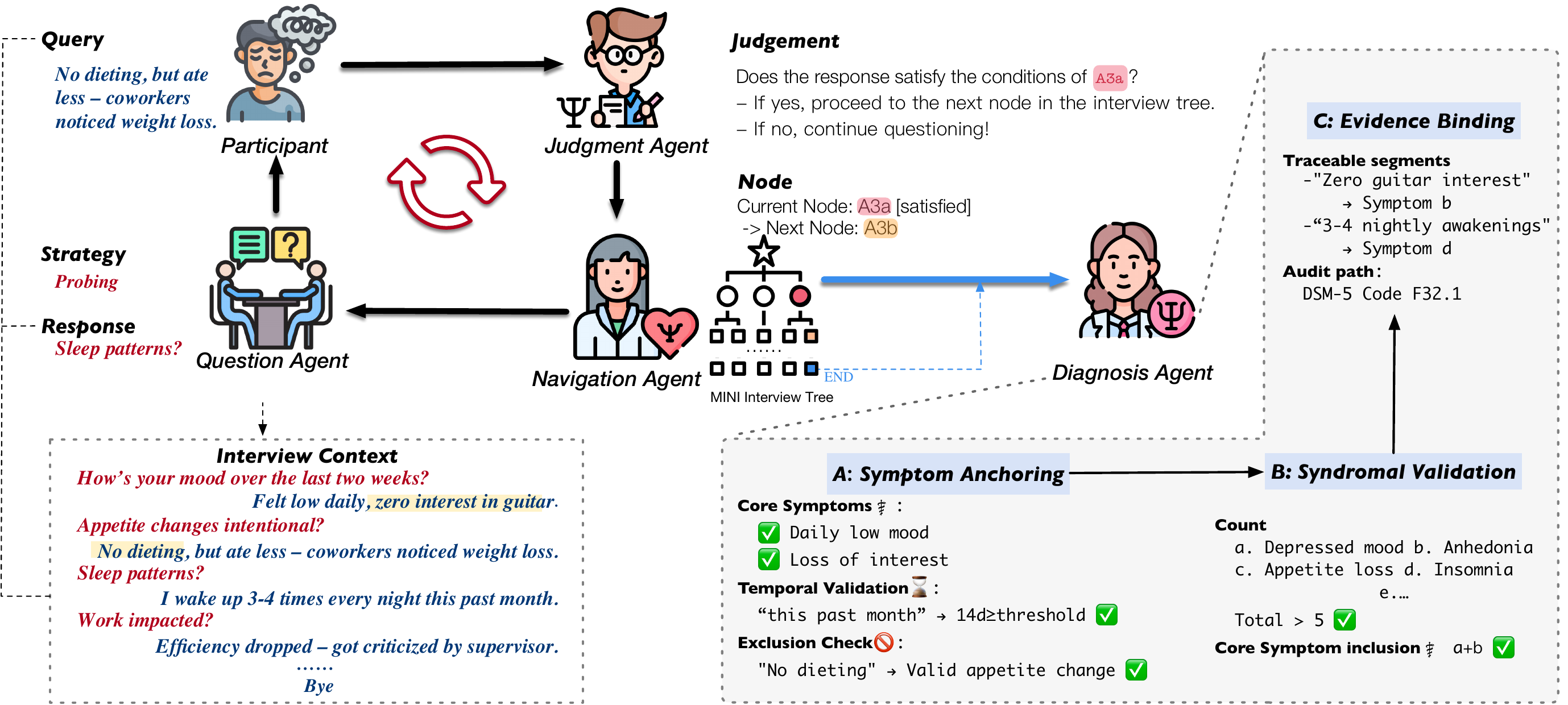} 
    \caption{Overview of \method framework. The framework consists of a navigation agent for interview management, a question agent for dynamic utterance generation, a judgment agent for symptom validity analysis, and a diagnosis agent for DSM-5 compliant conclusions, ensuring adherence to psychiatric protocols and conversational adaptability.}
    \label{fig-framework}
    % \vspace{-1.5em} 
\end{figure*}

\subsection{MINI: A Structured Clinical Interview}
The Mini International Neuropsychiatric Interview (MINI) is a structured diagnostic interview widely used for assessing major psychiatric disorders \citep{Sheehan1998TheMN}. Aligned with DSM-5 criteria \citep{apa2013dsm}, it provides a brief yet comprehensive assessment format that ensures consistency across different interviewers \citep{Lecrubier1997TheMI}.

The MINI implements a decision tree logic where patient responses guide the interview progression \citep{Amorim2019DiagnosticAO}, ensuring standardized administration while maintaining comprehensive evaluation \citep{Pettersson2018TheMI}. Notable for its efficiency, the interview can be completed in 15-30 minutes without compromising diagnostic accuracy \citep{Sheehan1998ComparativeMI}.
The instrument's reliability and validity have been validated across diverse populations \citep{Pinninti2003TheMI,sheehan2010reliability}, establishing it as a psychiatric evaluation standard \citep{vanVliet2015ValidationOT}. Its standardized structure facilitates research comparability \citep{Lecrubier1997TheMI} and presents opportunities for computational implementation \citep{Malhotra2015ComputerizedDA}. Our work adapts the MINI's framework into a multi-agent collaborative system while preserving its diagnostic rigor in an automated format.

\section{Methodology}
\label{sec:methodology}

% \begin{figure*}[t] 
%     \centering
%     \includegraphics[width=\linewidth]{figures/psycot.pdf} 
%     \caption{Overview of PsyCoT framework.}
%     \label{fig-psycot}
%     % \vspace{-1.5em} 
% \end{figure*}

% \subsection{Overview}
\method transforms the MINI interview into a collaborative multi-agent process, where four specialized components interact through clinically constrained data flows. 
The navigation agent (\S\ref{sec:navigation}) governs interview progression by enforcing MINI's branching logic, while the question agent (\S\ref{sec:question}) dynamically generates utterances that balance diagnostic probing with participant support. Participant responses are analyzed by the judgment agent (\S\ref{sec:judgment}) to determine symptom validity, with accumulated evidence processed by the diagnosis agent (\S\ref{sec:diagnosis}) to generate DSM-5 compliant conclusions through structured reasoning traces. This architecture ensures adherence to psychiatric assessment protocols while maintaining conversational adaptability.

\subsection{Navigation Agent}
\label{sec:navigation}
The navigation agent maintains overall control of the dialogue, enforcing strict adherence to the MINI’s semi-structured format. 
% In practice, it functions as an event-driven finite-state machine whose transitions are dictated by the interview’s branching rules. 
In practice, the agent governs dialog progression through hierarchical traversal of the MINI's interview tree, where node transitions follow strict branching logic derived from diagnostic protocols.
For instance, if a participant reports any sign of suicidal ideation, the agent routes the conversation to a deeper self-harm risk assessment node, whereas negative indications revert the session to a broader screening path. 
This design ensures that crucial items cannot be skipped or prematurely terminated: even if a participant attempts to change the topic, the agent compels the conversation to remain on the appropriate node until the relevant question set is satisfactorily answered. 

By separating flow control from language generation, the system prevents “hallucinations” or creative detours that might undermine diagnostic integrity. 
This agent functions as a management center for \method. 
It coordinates with the judgment agent to force-choice constraints, ensuring that critical responses (e.g., indicators of suicidal thoughts or mania) are explicitly addressed. 
When incoming judgment indicate that a participant's reply is ambiguous or shows reluctance to disclose information, the navigation agent automatically directs the question agent to rephrase or clarify queries. 
This process upholds the core MINI principle that certain inquiries must not be bypassed or left unaddressed.

\subsection{Question Agent}
\label{sec:question}
The question agent diversifies question phrasing to sustain engagement while adhering to diagnostic intent.
For symptom assessment, it rephrases clinical constructs into conversational probes without deviating from diagnostic intent. 
For example, when probing for anhedonia — a core depression symptom — it adapts wording based on conversation history: a participant mentioning hobbies might hear "Have your favorite activities felt less rewarding recently?", while someone discussing social habits could receive "Do you still look forward to spending time with friends like before?". This contextual rephrasing prevents robotic repetition without compromising screening accuracy.
When the LLM detects confusion through contextual inconsistencies (e.g., conflating chronic anxiety with momentary stress), it activates explanations: “Earlier you mentioned constant worry about work. This daily pattern aligns with what we call generalized anxiety.”. 
Real-time dialogue analysis further identifies emotional distress cues through lexical patterns (e.g., repeated “overwhelmed”, abrupt topic shifts), triggering empathetic support that maintains clinical progression: “I appreciate you sharing this. To help clarify, how many days last week did this loneliness feel most intense?”. 
Strategy shifts emerge organically from LLM-powered context parsing rather than rigid hierarchies, enabling fluid transitions between diagnostic precision and human-centered responsiveness.

\subsection{Judgment Agent}
\label{sec:judgment}
This agent operates real-time alignment between natural dialogue and structured diagnostic evaluation. 
It continuously evaluates response adequacy against predefined MINI criteria through three decision thresholds: 
direct matching of operationalized definitions (e.g., interpreting "I wake up panicking daily" as meeting generalized anxiety's physiological symptom threshold), 
semantic comprehension of clinically equivalent expressions ("constant worrying" → generalized anxiety criteria), 
and ambiguity resolution after 5 unproductive turns. 
Unmet criteria trigger recursive clarification requests through the question agent, while confirmed matches authorize state transitions via the navigation agent. The forced-choice mechanism activates upon persistent ambiguity, presenting binary MINI-compliant options ("Would you describe this as [exact phrasing from MINI] or [alternative]?") to maintain diagnostic fidelity. This dual-layer validation - combining dynamic interpretation with protocol enforcement - ensures structured evidence collection while preserving natural dialogue flow.

\subsection{Diagnosis Agent} 
\label{sec:diagnosis}
The diagnosis agent operationalizes the PsyCoT (Psychometric Chain-of-Thought) framework to synthesize clinical evidence into DSM-5 compliant diagnoses through three interpretable reasoning phases.

At the \textbf{symptom anchoring} stage, each diagnostic criterion (e.g., "$\geq$ 5 symptoms over two weeks" for depression, "pervasive worry for six months" for generalized anxiety) is translated into multi-step validation logic. This includes symptom existence confirmation,temporal verification,  and exclusionary condition checks (e.g., ruling out medical causes). The process transforms textual criteria into auditable decision paths. 
\textbf{Diagnostic synthesis} then performs syndromal validation by confirming that the symptom profile satisfies DSM-5 requirements. This includes verifying that the total number of distinct symptoms exceeds the diagnostic threshold, that core symptoms.
% are present, and that all symptoms are anchored to appropriate durations and not attributable to exclusionary factors. 
This structured validation ensures that both the quantity and quality of symptoms align with the formal criteria before proceeding to diagnosis.
% Severity stratification adapts dynamically to each disorder's criteria without external scales.
Finally, \textbf{evidence binding} creates transparent audit trails by linking diagnostic conclusions to originating dialogue evidence and verification steps. Each judgment documents: which responses confirmed criteria fulfillment, how duration/exclusion rules were applied, and uncertainty sources requiring clinical review.

This three-phase PsyCoT implementation achieves transparency while maintaining MINI's diagnostic rigor, enabling clinicians to trace how conversational fragments map to specific DSM-5 codes through the reasoning chain.

\section{Experiments}
\paragraph{Data Collection}
We evaluated \method on 1,002 high-quality interview sessions collected through university mental health services. The interface for data collection is displayed in the Appendix~\ref{app:demo}. 
Considering ethical constraints, we first construct simulated participants with different mental illnesses based on LLMs. Then, we let various interviewers engage in conversations with them, and finally, we evaluate the interviews to compare the performance of different interviewers.
Diagnostic annotations were performed by two licensed psychologists from APA-accredited university clinics, with 10-15 years of clinical experience. Following institutional compensation standards of \$100/hour, these experts implemented a three-phase annotation protocol. First, they jointly annotated 150 cases, achieving strong inter-rater reliability with an ICC of 0.78 and mean symptom dimension agreement of 0.69. They then independently annotated the remaining cases with weekly calibration sessions, maintaining high agreement (Cohen's $\kappa$ > 0.85). A final random re-sampling of 50 cases confirmed the annotation quality with an ICC of 0.87.

\input{tables/human_result}

\subsection{Dialogue Quality Evaluation}
\subsubsection{Experimental Setup}

\paragraph{Baselines} We evaluate four baseline approaches for psychiatric interview simulation:
(1) \textit{Direct}: The LLM follows explicit instructions to conduct diagnostic interviews without role-specific prompting or domain knowledge integration.
(2) \textit{Role-Play}: The LLM is instructed to simulate clinician expertise through the prompt "Act as a board-certified psychiatrist" while maintaining standard conversational capabilities.
(3) \textit{Knowledge}: The model is enhanced with DSM-5 diagnostic criteria and psychopathology knowledge graphs, enabling disorder-specific symptom evaluation during dialogues.
(4) \textit{MINI}: The LLM implements the MINI decision-tree architecture through constrained response generation and diagnostic branching logic.

\paragraph{Metrics} For evaluation, experts rated each dialogue on 5-point Likert scales for:
(1) \texttt{Relevance}: Measures response alignment with participant's queries and emotional context;
(2) \texttt{Accuracy}: Evaluates psychological principle adherence and emotion judgment correctness; 
(3) \texttt{Completeness}: Assesses comprehensive coverage of critical issues with logical structure; 
(4) \texttt{Guidance}: Examines proactive conversation steering and emotional exploration facilitation.
The detailed scale guidebook is given in Appendix~\ref{app:scale}.

\input{tables/main_result}

\subsubsection{Results}
The evaluation results in Table~\ref{tab:human_results} demonstrate that \method achieves superior performance in almost all key dimensions. 
In relevance, our multi-agent architecture ensures systematic adherence to MINI nodes through the interview tree in the navigation agent, dynamically filtering off-topic responses while maintaining natural transitions between diagnostic topics. 
In completeness, the judgment agent's forced conclusive symptom matching and the navigation agent's state-aware questioning strategy work synergistically to exhaustively cover required diagnostic criteria, addressing the incompleteness of rigid MINI-based baselines. 
The guidance score benefits from the questioning agent's hybrid strategy where empathetic feedback stabilizes participant engagement. 
Although \method slightly trails the knowledge baseline in accuracy, this marginal gap likely stems from our intentional trade-off between strict symptom classification and participant-centered adaptability.
The judgment agent occasionally prioritizes participant comfort over forcing definitive answers, whereas the knowledge baseline's static disease definitions enable more deterministic scoring. 
Crucially, \method's diagnostic agent compensates for this through DSM-5-aligned reasoning chains, ensuring clinically valid interpretations despite minor accuracy variations in individual responses. These results collectively validate that \method's orchestration of structured protocol adherence, adaptive interaction, and expert-knowledge integration achieves human-like competency in psychological interviews.
Detailed results can be found in the Appendix~\ref{sec:app-zeroshot}.

\subsection{Diagnostic Performance Evaluation}

\subsubsection{Experimental Setup}
\paragraph{Baselines}We evaluate diagnostic results on 1,002 expert-annotated samples using four LLMs: \texttt{GPT-4o}, \texttt{Claude-3.5-sonnet}, \texttt{GLM-Zero}, and \texttt{DeepSeek-R1}, with temperature set to 0.7 for all models.
Each model runs in three configurations:
(1) Vanilla: End-to-end prediction from transcripts; 
(2) CoT: Free-form reasoning before diagnosis.
(c) PsyCoT: Our DSM-5-aligned structured reasoning.% chain.

\paragraph{Metrics} Following previous studies \cite{burdisso2023node, Chen2024DepressionDI}, we evaluate model performance using: (1) macro-F1 within case/control groups, (2) overall macro-F1 and accuracy across all data, and (3) Cohen's $\kappa$ for agreement with expert annotations.

\subsubsection{Results}
The evaluation in Table~\ref{tab:main_results} demonstrates PsyCoT's effectiveness in enhancing LLM-based psychiatric assessment across diagnostic complexity levels. 
The framework consistently outperforms baseline methods, particularly in high-stakes scenarios where clinical precision matters most. 
For suicide risk detection, a task requiring nuanced differentiation between transient distress and pathological patterns, PsyCoT elevates $\kappa$ to 0.839-0.942, achieving clinical acceptability with $\kappa$ exceeding 0.6, while baseline methods fall short with values between 0.259 and 0.427. 
This improvement stems from PsyCoT's symptom anchoring mechanism, which maps symptom mentions to DSM-5 thresholds while filtering confounding factors.
PsyCoT also mitigates model capability disparities: smaller models like GLM-Zero approach GPT-4o's performance in depression screening with Macro-F1 scores of 0.772 versus 0.767 when guided by structured clinical reasoning. 
The framework balances sensitivity-specificity tradeoffs by maintaining control group F1 above 0.968 while doubling case detection rates for social anxiety disorders. This reflects its syndromic validation phase, which evaluates symptom interdependencies to reject isolated signals, such as distinguishing pathological avoidance from situational nervousness.
% In comorbidity-prone conditions like generalized anxiety, PsyCoT’s severity contextualization adds granularity, quantifying impairment levels where vanilla methods achieve surface-level accuracy. 
The multi-agent architecture preserves MINI's procedural rigor, as evidenced by DeepSeek-R1's improved adherence to diagnostic hierarchies under PsyCoT, reducing misclassification of subclinical cases. These results suggest that encoding domain-specific reasoning chains compensates for LLMs' tendency toward heuristic judgments, bridging the gap between open-ended language understanding and standardized diagnostic protocols.

\section{Discussion and Analysis}

\subsection{The Effect of Few-shot Learning}
Results in Table \ref{tab:few-shot} demonstrate the superior performance of PsyCoT across four psychological disorders. In the zero-shot setting, PsyCoT achieves substantial improvements over baseline methods, particularly in suicide risk assessment with F1 scores of 0.945 and generalized anxiety detection with F1 scores of 0.946, indicating robust integration of clinical knowledge. The framework maintains consistent performance across disorders, with F1 scores ranging from 0.803 to 0.946 in zero-shot settings, suggesting effective incorporation of DSM-5 criteria and MINI protocols. While all methods benefit from few-shot learning, PsyCoT exhibits more stable improvements ranging from 1.5\% to 5.9\%, compared to the larger variations seen in baseline approaches—Vanilla methods show improvements from 2.7\% to 26.3\%, while CoT methods show improvements from 1.7\% to 20.9\%. This stability is particularly evident in challenging tasks such as depression assessment, where PsyCoT achieves a significant improvement from baseline with F1 increasing from 0.679 to 0.850. The most notable performance gain is observed in social anxiety detection, where the baseline method shows substantial few-shot improvement of 26.3\%, indicating the complexity of social anxiety manifestations. These results validate our multi-agent architecture's effectiveness in conducting structured psychological interviews, with PsyCoT demonstrating robust performance across different clinical scenarios, particularly in critical tasks such as suicide risk assessment with few-shot F1 reaching 0.963. The framework's consistent performance across disorders and minimal reliance on few-shot examples suggests successful integration of clinical expertise into the reasoning process, supporting its potential for broad clinical application.
Detailed few-shot methods results can be found in the Appendix~\ref{sec:app-fewshot}.

\input{tables/fewshot}
\begin{figure*}[t] 
    \centering
    \includegraphics[width=0.99\linewidth]{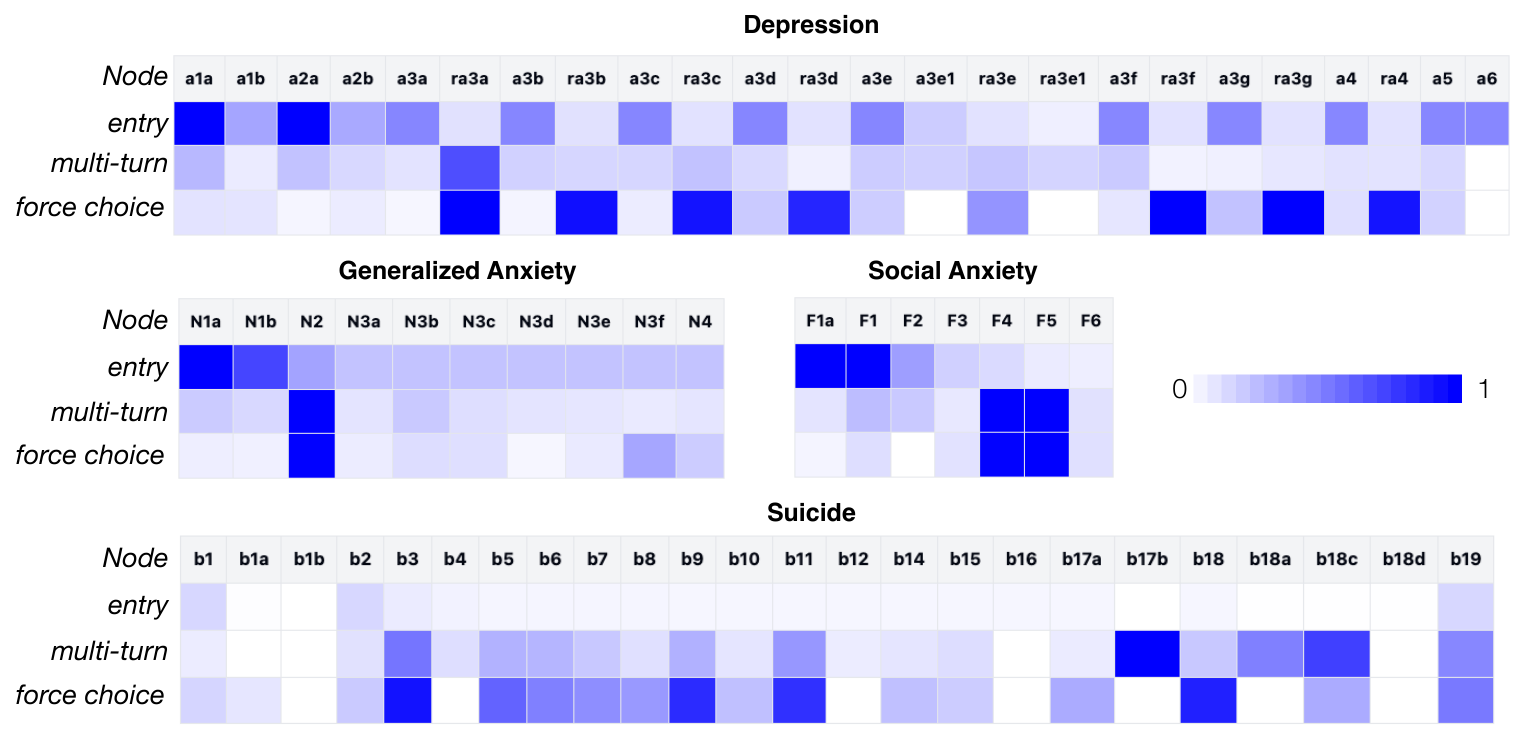} 
    \caption{Heatmap illustrating participant interaction patterns with interview nodes across different disorders, with darker blue indicating higher percentage at different assessment nodes.}
    \label{fig:heatmap}
    % \vspace{-1.5em} 
\end{figure*}

\subsection{Interview Tree Nodes Analysis}

Figure~\ref{fig:heatmap} presents heatmaps visualizing participant interaction patterns with different interview nodes. The color intensity represents normalized metrics ranging from 0 to 1.
In the depression assessment, participants showed high interaction intensity at entry nodes \texttt{a1a} and \texttt{a2a}, while multi-turn interactions dominated middle-phase nodes such as \texttt{ra3a}, and force choice questions were prominent in later stages including \texttt{ra3g} and \texttt{ra4}. This progression suggests a structured approach from open dialogue to specific symptom confirmation. 
The generalized anxiety assessment demonstrated concentrated interaction at early nodes \texttt{N1a} and \texttt{N2} with both entry and multi-turn patterns, indicating the importance of establishing rapport before detailed symptom exploration. 
Social anxiety evaluation exhibited a unique pattern where entry interactions peaked at \texttt{F1a} and \texttt{F1}, followed by increased force choice and multi-turn interactions at nodes \texttt{F4} through \texttt{F5}, highlighting the necessity of trust-building before exploring specific social scenarios. 
Suicide tendency assessment displayed more dispersed interaction patterns, with heightened intensity at nodes \texttt{b17a} and \texttt{b18c}, reflecting the complexity and sensitivity of this evaluation domain. 
The varying intensity patterns across different nodes and interaction types suggest that effective mental health assessment requires a dynamic approach, adapting the interaction style based on both the specific dimension being evaluated and the stage of assessment.

\subsection{Case Study: Explainable Diagnosis}

As shown in Table \ref{tab-case}, the generalized anxiety case demonstrates \method's capacity to transform ambiguous dialog exchanges into clinically valid diagnoses through transparent PsyCoT reasoning.

\begin{table}[ht]
\centering
\small
\renewcommand\arraystretch{1.2}
\begin{tabular}{|p{0.93\linewidth}|}
\hline
\textbf{Case ID:} GAD-xxxxxxxx \\
\textbf{Diagnosis:} Generalized Anxiety (F41.1) \\
\hline
\textbf{[Criterion A]} Excessive worry confirmed via multi-turn temporal validation: \newline
\textit{...occur every few days} $\rightarrow$ \textbf{meets 6-month persistence through cumulative frequency analysis} \\ \\

\textbf{[Criterion B]} Uncontrollability established through coping efficacy assessment: \newline
\textit{I find it very difficult to control my anxiety} $\rightarrow$ \textbf{despite self-help attempts} \\ \\

\textbf{[Criterion C]} Physiological symptoms mapped through clinical equivalence: \newline
\textit{Stomach spasms} $\rightarrow$ \textbf{muscle tension} \newline
\textit{Sleep disturbance weighted as secondary evidence:} $\rightarrow$ \textbf{frequent nightmares with fatigue} \\ \\

\textbf{[Criterion E]} Functional impairment validated: \newline
\textit{Inappropriate attitudes during communication} $\rightarrow$ \textbf{social domain impact} \\
\hline
\end{tabular}
\caption{Case study for PsyCoT. \textit{Italic} words are the original dialogue from the interview. \textbf{Bolden} words are the analyzed symptoms.}
\label{tab-case}
\end{table}

When the participant initially described intermittent worry patterns (\textit{occur every few days}), the agent clarified temporal parameters through multi-turn verification, mathematically aggregating frequency reports to confirm DSM-5's 6-month duration threshold. 
Core symptom validation involved distinguishing pathological uncontrollability (\textit{find it very difficult to control my anxiety} despite self-help attempts) from transient stress through semantic decomposition of coping efficacy descriptions. Physiological markers were systematically mapped to diagnostic criteria: self-reported \textit{stomach spasms} validated muscle tension through visceromotor equivalence reasoning, while sleep quality complaints (\textit{frequent nightmares and daytime fatigue}) were weighted as secondary evidence under polythetic rules. Crucially, the agent maintained an audit trail linking each diagnostic conclusion to 11 dialog segments - for instance, binding social impairment to specific interpersonal conflict descriptions (\textit{inappropriate attitudes during communication}) while excluding alternative explanations through negative medical history screening. This end-to-end traceability, where conversational fragments flow through symptom anchoring (temporal verification), diagnostic synthesis (symptom constellation analysis), and evidence binding (DSM-5 criteria mapping), exemplifies how PsyCoT bridges natural language understanding with clinical decision protocols. 
Due to space limitations, we only show the most core part of the reasoning.

\section{Conclusion}
This work addresses the critical challenge of aligning LLM-powered mental health assessment with clinical diagnostic standards through \method, the first multi-agent framework that operationalizes structured psychiatric interviews. By translating the MINI protocol into dynamic computational workflows, our framework overcomes three fundamental limitations of existing approaches: (1) systematic adherence to diagnostic logic through navigation and judgment agents ensures clinical validity, (2) adaptive questioning strategies enable human-like conversational flexibility while maintaining assessment integrity, and (3) the novel PsyCoT mechanism provides transparent diagnostic reasoning that maps symptoms to DSM-5 criteria through intermediate psychiatric constructs. Extensive evaluation across 1,002 real-world cases demonstrates \method's superior diagnostic agreement with expert clinicians while maintaining natural interaction flow—a 32\% improvement over single-agent LLM baselines.

\section*{Ethical Considerations}

\paragraph{Ethical Review and Informed Consent}
This study rigorously adheres to ethical standards through comprehensive measures spanning ethical oversight, psychological safeguards, and data governance. The experimental protocol received formal approval from an institutional review board. Prior to participation, all 1,002 individuals provided informed consent, explicitly acknowledging the non-diagnostic nature of the research, their unconditional right to withdraw without consequences, and the exclusive use of anonymized data for institutional management and academic publication. No biometric identifiers were collected or stored.
\paragraph{Psychological Risk Mitigation}
To address psychological risks, a multi-tiered intervention protocol was established for participants exhibiting signs of depression, anxiety, or suicidal ideation. Immediate referrals to certified psychological hotlines were provided during initial assessments, followed by secondary evaluations conducted by licensed clinicians within 72 hours. High-risk cases involving self-harm or suicidal intent triggered collaborative alerts between researchers and institutional authorities, with all procedures reviewed and validated by a panel of clinical psychology experts to align with the Ethical Code for Clinical and Counseling Psychology issued by the Chinese Psychological Society.
\paragraph{Data Management and AI Boundaries}
Data security was ensured through storage on a government-certified cloud platform meeting cybersecurity standards, accessible only to authorized research members. Findings were aggregated into non-identifiable group reports to preclude individual disclosure. Notably, the study enforced strict boundaries on AI applications: no AI-generated analyses were directly shared with participants, clinical diagnoses were reserved exclusively for qualified psychiatrists, and no data were transferred to third-party commercial entities. These safeguards collectively prioritize participant welfare, data integrity, and compliance with regional ethical-legal frameworks.

\section*{Limitations and Future Work}
The current implementation of \method faces several important constraints. While our evaluation demonstrates strong agreement with expert judgments, longitudinal studies with diverse participant populations are needed to validate \method's reliability in clinical settings. Technical aspects requiring further development include enhancement of the PsyCoT reasoning framework for complex comorbidity patterns and improvement of real-time emotion detection capabilities. Practical deployment considerations encompass clinical integration protocols, healthcare provider training requirements, and robust data privacy frameworks. These challenges, while significant, also present opportunities for advancing AI-assisted mental healthcare through careful attention to clinical, ethical, and technical considerations.

\bibliography{acl_latex}

\appendix

\input{Appendix/zeroshot_results}
\input{Appendix/fewshot_results}

\input{Appendix/demo}

\input{Appendix/scale}
\input{Appendix/prompt}

\end{document}

%% file: tables/human_result.tex
\begin{table}
\small
  \centering
  \resizebox{0.48\textwidth}{!}{
  \begin{tabular}{lccccc}
    \toprule
    \textbf{Method} & \textbf{Rel.} & \textbf{Acc.} & \textbf{Comp.} & \textbf{Gui.}\\
    \midrule
    Direct & 3.506 & 3.463 & 3.350 & 3.294 \\ \midrule
    Role-Play & 3.438 & 3.438 & 3.331 & 3.244 \\ \midrule
    Knowledge & 3.544 & \textbf{3.506} & 3.425 & 3.325 \\ \midrule
    MINI & 3.519 & 3.475 & 3.369 & 3.313 \\ \midrule 
    % \midrule
    \textbf{\method} & \textbf{3.576} & 3.503 & \textbf{3.497 }& \textbf{3.460} \\ 
    % Human & & & & \\
    \bottomrule
  \end{tabular}}
    \caption{Results of dialogue quality evaluation. Rel.: Relevance, Acc.: Accuracy, Comp.: Completeness, Gui.: Guidance.} 
  \label{tab:human_results}
\end{table}

%% file: tables/main_result.tex
\begin{table*}[htbp]
\centering

\renewcommand\arraystretch{1.25}
\resizebox{\textwidth}{!}{
\begin{tabular}{lccccccccccccc}
\toprule
\multirow{2}{*}{\textbf{Disorder}} & \multirow{2}{*}{\textbf{Metric}} & \multicolumn{3}{c}{GPT-4o} & \multicolumn{3}{c}{Claude-3.5-sonnet} & \multicolumn{3}{c}{GLM-Zero} & \multicolumn{3}{c}{DeepSeek-R1} \\
\cmidrule(lr){3-5} \cmidrule(lr){6-8} \cmidrule(lr){9-11} \cmidrule(lr){12-14}
 & & Vanilla & CoT & PsyCoT & Vanilla & CoT & PsyCoT & Vanilla & CoT & PsyCoT & Vanilla & CoT & PsyCoT \\
\midrule
\multirow{5}{*}{Depression}
 & Control F1 & 0.930 & 0.938 & \sethlcolor{soulblue}\hl{\textbf{0.976}} & 0.925 & 0.925 & \sethlcolor{soulblue}\hl{\textbf{0.978}} & 0.877 & 0.887 & \sethlcolor{soulblue}\hl{\textbf{0.968}} & 0.912 & 0.914 & \sethlcolor{soulblue}\hl{\textbf{0.977}} \\
 & Case F1 & 0.396 & 0.429 & \sethlcolor{soulblue}\hl{\textbf{0.558}} & 0.393 & 0.396 & \sethlcolor{soulblue}\hl{\textbf{0.632}} & 0.333 & 0.356 & \sethlcolor{soulblue}\hl{\textbf{0.576}} & 0.408 & 0.432 & \sethlcolor{soulblue}\hl{\textbf{0.639}} \\
 & Macro-F1 & 0.663 & 0.683 & \sethlcolor{soulblue}\hl{\textbf{0.767}} & 0.659 & 0.661 & \sethlcolor{soulblue}\hl{\textbf{0.805}} & 0.605 & 0.622 & \sethlcolor{soulblue}\hl{\textbf{0.772}} & 0.660 & 0.673 & \sethlcolor{soulblue}\hl{\textbf{0.808}} \\
 & Accuracy & 0.875 & 0.888 & \sethlcolor{soulblue}\hl{\textbf{0.954}} & 0.867 & 0.866 & \sethlcolor{soulblue}\hl{\textbf{0.958}} & 0.792 & 0.808 & \sethlcolor{soulblue}\hl{\textbf{0.941}} & 0.846 & 0.850 & \sethlcolor{soulblue}\hl{\textbf{0.957}} \\
 & $\kappa$ & 0.337 & 0.374 & \sethlcolor{soulblue}\hl{\textbf{0.535}} & 0.331 & 0.335 & \sethlcolor{soulblue}\hl{\textbf{0.610}} & 0.256 & 0.282 & \sethlcolor{soulblue}\hl{\textbf{0.544}} & 0.343 & 0.370 & \sethlcolor{soulblue}\hl{\textbf{0.616}} \\

\midrule
\multirow{5}{*}{Generalized Anxiety}
 & Control F1 & \sethlcolor{soulyellow}\hl{\textbf{0.974}} & \sethlcolor{soulyellow}\hl{\textbf{0.974}} & 0.973 & 0.970 & 0.972 & \sethlcolor{soulyellow}\hl{\textbf{0.980}} & 0.931 & 0.939 & \sethlcolor{soulyellow}\hl{\textbf{0.985}} & 0.969 & 0.974 & \sethlcolor{soulyellow}\hl{\textbf{0.990}} \\
 & Case F1 & 0.866 & \sethlcolor{soulyellow}\hl{\textbf{0.868}} & 0.857 & 0.845 & 0.860 & \sethlcolor{soulyellow}\hl{\textbf{0.899}} & 0.756 & 0.778 & \sethlcolor{soulyellow}\hl{\textbf{0.929}} & 0.851 & 0.872 & \sethlcolor{soulyellow}\hl{\textbf{0.952}} \\
 & Macro-F1 & 0.920 & \sethlcolor{soulyellow}\hl{\textbf{0.921}} & 0.915 & 0.907 & 0.916 & \sethlcolor{soulyellow}\hl{\textbf{0.940}} & 0.843 & 0.858 & \sethlcolor{soulyellow}\hl{\textbf{0.957}} & 0.910 & 0.923 & \sethlcolor{soulyellow}\hl{\textbf{0.971}} \\
 & Accuracy & \sethlcolor{soulyellow}\hl{\textbf{0.957}} & \sethlcolor{soulyellow}\hl{\textbf{0.957}} & 0.955 & 0.949 & 0.953 & \sethlcolor{soulyellow}\hl{\textbf{0.967}} & 0.892 & 0.904 & \sethlcolor{soulyellow}\hl{\textbf{0.975}} & 0.949 & 0.957 & \sethlcolor{soulyellow}\hl{\textbf{0.983}} \\
 & $\kappa$ & 0.841 & \sethlcolor{soulyellow}\hl{\textbf{0.842}} & 0.831 & 0.815 & 0.832 & \sethlcolor{soulyellow}\hl{\textbf{0.880}} & 0.690 & 0.719 & \sethlcolor{soulyellow}\hl{\textbf{0.914}} & 0.821 & 0.847 & \sethlcolor{soulyellow}\hl{\textbf{0.941}} \\

\midrule
\multirow{5}{*}{Social Anxiety}
 & Control F1 & \sethlcolor{soulorange}\hl{\textbf{0.984}} & 0.982 & 0.992 & 0.974 & 0.976 & \sethlcolor{soulorange}\hl{\textbf{0.991}} & 0.948 & 0.951 & \sethlcolor{soulorange}\hl{\textbf{0.995}} & 0.974 & 0.974 & \sethlcolor{soulorange}\hl{\textbf{0.996}} \\
 & Case F1 & 0.752 & 0.736 & \sethlcolor{soulorange}\hl{\textbf{0.800}} & 0.667 & 0.686 & \sethlcolor{soulorange}\hl{\textbf{0.786}} & 0.505 & 0.519 & \sethlcolor{soulorange}\hl{\textbf{0.898}} & 0.667 & 0.667 & \sethlcolor{soulorange}\hl{\textbf{0.918}} \\
 & Macro-F1 & 0.868 & 0.859 & \sethlcolor{soulorange}\hl{\textbf{0.896}} & 0.820 & 0.831 & \sethlcolor{soulorange}\hl{\textbf{0.888}} & 0.727 & 0.735 & \sethlcolor{soulorange}\hl{\textbf{0.946}} & 0.820 & 0.820 & \sethlcolor{soulorange}\hl{\textbf{0.957}} \\
 & Accuracy & 0.969 & 0.967 & \sethlcolor{soulorange}\hl{\textbf{0.984}} & 0.952 & 0.956 & \sethlcolor{soulorange}\hl{\textbf{0.982}} & 0.906 & 0.911 & \sethlcolor{soulorange}\hl{\textbf{0.990}} & 0.952 & 0.952 & \sethlcolor{soulorange}\hl{\textbf{0.992}} \\
 & $\kappa$ & 0.736 & 0.719 & \sethlcolor{soulorange}\hl{\textbf{0.792}} & 0.644 & 0.665 & \sethlcolor{soulorange}\hl{\textbf{0.777}} & 0.467 & 0.482 & \sethlcolor{soulorange}\hl{\textbf{0.893}} & 0.644 & 0.644 & \sethlcolor{soulorange}\hl{\textbf{0.914}} \\

\midrule
\multirow{5}{*}{Suicide Risk}
 & Control F1 & 0.965 & 0.963 & \sethlcolor{soulgreen}\hl{\textbf{0.995}} & 0.963 & 0.966 & \sethlcolor{soulgreen}\hl{\textbf{0.988}} & 0.941 & 0.937 & \sethlcolor{soulgreen}\hl{\textbf{0.995}} & 0.965 & 0.966 & \sethlcolor{soulgreen}\hl{\textbf{0.995}} \\
 & Case F1 & 0.385 & 0.340 & \sethlcolor{soulgreen}\hl{\textbf{0.942}} & 0.430 & 0.454 & \sethlcolor{soulgreen}\hl{\textbf{0.852}} & 0.316 & 0.326 & \sethlcolor{soulgreen}\hl{\textbf{0.944}} & 0.396 & 0.425 & \sethlcolor{soulgreen}\hl{\textbf{0.947}} \\
 & Macro-F1 & 0.675 & 0.651 & \sethlcolor{soulgreen}\hl{\textbf{0.968}} & 0.697 & 0.710 & \sethlcolor{soulgreen}\hl{\textbf{0.920}} & 0.629 & 0.631 & \sethlcolor{soulgreen}\hl{\textbf{0.969}} & 0.681 & 0.695 & \sethlcolor{soulgreen}\hl{\textbf{0.971}} \\
 & Accuracy & 0.933 & 0.930 & \sethlcolor{soulgreen}\hl{\textbf{0.990}} & 0.931 & 0.935 & \sethlcolor{soulgreen}\hl{\textbf{0.977}} & 0.892 & 0.884 & \sethlcolor{soulgreen}\hl{\textbf{0.990}} & 0.933 & 0.935 & \sethlcolor{soulgreen}\hl{\textbf{0.991}} \\
 & $\kappa$ & 0.363 & 0.318 & \sethlcolor{soulgreen}\hl{\textbf{0.936}} & 0.400 & 0.427 & \sethlcolor{soulgreen}\hl{\textbf{0.839}} & 0.259 & 0.262 & \sethlcolor{soulgreen}\hl{\textbf{0.938}} & 0.373 & 0.401 & \sethlcolor{soulgreen}\hl{\textbf{0.942}} \\
\bottomrule
\end{tabular}}
\caption{Diagnostic performance evaluation results. PsyCoT is our proposed method. Vanilla represents end-to-end prediction from transcripts. CoT is free-form reasoning before diagnosis.}
\label{tab:main_results}
\end{table*}

%% file: tables/fewshot.tex
\begin{table}[htbp]
\renewcommand\arraystretch{1.2}
\centering
\resizebox{0.48\textwidth}{!}{
\begin{tabular}{llccc}
\toprule
\multicolumn{2}{c}{\multirow{2}{*}{Disorder \& Method}} & \multicolumn{3}{c}{Macro-F1} \\
\cmidrule(lr){3-5}
\multicolumn{2}{c}{} & Zero-Shot & Few-Shot & $\Delta$ (\%) \\
\midrule
\multirow{3}{*}{Depression} 
 & Vanilla & 0.679 & 0.723 & +6.5 \\
 & CoT & 0.700 & 0.755 & +7.9 \\
 & PsyCoT & 0.803 & \textbf{0.850} & +5.9 \\
% \addlinespace
\midrule
\multirow{3}{*}{Generalized Anxiety} 
 & Vanilla & 0.895 & 0.919 & +2.7 \\
 & CoT & 0.904 & 0.919 & +1.7 \\
 & PsyCoT & 0.946 & \textbf{0.960} & +1.5 \\
% \addlinespace
 \midrule
\multirow{3}{*}{Social Anxiety} 
 & Vanilla & 0.729 & 0.921 & +26.3 \\
 & CoT & 0.761 & 0.920 & +20.9 \\
 & PsyCoT & 0.927 & \textbf{0.942} & +1.6 \\
% \addlinespace
\midrule
\multirow{3}{*}{Suicide} 
 & Vanilla & 0.763 & 0.815 & +6.8 \\
 & CoT & 0.775 & 0.810 & +4.5 \\
 & PsyCoT & 0.945 & \textbf{0.963} & +1.9 \\
\bottomrule
\end{tabular}
}
\caption{Few-shot performance improvements across disorders with comparison of methods (Vanilla, CoT, PsyCoT). \textbf{Bold} values indicate best results from our proposed PsyCoT framework.}
\label{tab:few-shot}
\end{table}

%% file: Appendix/zeroshot_results.tex
\section{Detailed Diagnosis Results}
\subsection{Zero-shot Results}
\label{sec:app-zeroshot}

The zero-shot evaluation reveals significant performance variations across psychological disorders and prompting strategies, as shown in Table \ref{tab:zero-shot-main_results}. For depression diagnosis, PsyCoT consistently outperformed vanilla and chain-of-thought prompting, achieving Macro-F1 scores between 0.767 and 0.808 and Cohen's $\kappa$ values between 0.535 and 0.616 across models. GPT-4o with PsyCoT improved case F1 from 0.396 to 0.558, representing a 41\% increase compared to vanilla prompting and demonstrating enhanced sensitivity to minority-class identification. This pattern extends to high-stakes domains: for suicide risk assessment, PsyCoT increased GPT-4o's case F1 from 0.385 to 0.942, a 145\% improvement that substantially reduced false negatives. Model-specific disparities emerged in the results, with Claude-3.5 showing stronger anxiety detection with a Macro-F1 of 0.940 using PsyCoT, while GLM-Zero benefited most in social anxiety classification with Macro-F1 improving from 0.946 to 0.975, a 29\% increase. The $\kappa$ values between 0.936 and 0.942 for suicide risk suggest PsyCoT's improved clinical reliability by aligning with multi-rater consensus standards. A critical limitation persists: while control-group precision remains near-perfect with values between 0.962 and 0.997 across disorders, case recall fluctuates considerably, ranging from 0.439 to 0.966 for depression, indicating potential dataset bias toward majority classes. These results highlight task-specific optimization requirements and the necessity of domain-adapted prompting frameworks in clinical NLP applications.

\begin{table*}[htbp]
\centering
\resizebox{\textwidth}{!}{
\begin{tabular}{llllllllllll}
\toprule
\multirow{2}{*}{Disorder} & \multirow{2}{*}{Model} & \multirow{2}{*}{Method} & \multicolumn{3}{c}{Control} & \multicolumn{3}{c}{Case} & \multirow{2}{*}{Macro-F1} & \multirow{2}{*}{Acc.} & \multirow{2}{*}{$\kappa$} \\
\cmidrule(lr){4-6} \cmidrule(lr){7-9}
 & & & P & R & F1 & P & R & F1 & & & \\
\midrule
\multirow{12}{*}{Depression} 
 & \multirow{3}{*}{GPT-4o} & Vanilla & 0.971 & 0.893 & 0.930 & 0.291 & 0.621 & 0.396 & 0.663 & 0.875 & 0.337 \\
 & & CoT & \textbf{0.973} & 0.906 & 0.938 & 0.323 & \textbf{0.636} & 0.429 & 0.683 & 0.888 & 0.374 \\
 & & PsyCoT & 0.962 & \textbf{0.990} & \textbf{0.976} & \textbf{0.763} & 0.439 & \textbf{0.558} & \textbf{0.767} & \textbf{0.954} & \textbf{0.535} \\
\cmidrule(lr){2-12}
 & \multirow{3}{*}{Claude-3.5} & Vanilla & 0.973 & 0.882 & 0.925 & 0.281 & 0.652 & 0.393 & 0.659 & 0.867 & 0.331 \\
 & & CoT & \textbf{0.974} & 0.880 & 0.925 & 0.282 & \textbf{0.667} & 0.396 & 0.661 & 0.866 & 0.335 \\
 & & PsyCoT & 0.969 & \textbf{0.987} & \textbf{0.978} & \textbf{0.750} & 0.545 & \textbf{0.632} & \textbf{0.805} & \textbf{0.958} & \textbf{0.610} \\
\cmidrule(lr){2-12}
 & \multirow{3}{*}{GLM-Zero} & Vanilla & 0.981 & 0.793 & 0.877 & 0.211 & 0.788 & 0.333 & 0.605 & 0.792 & 0.256 \\
 & & CoT & \textbf{0.983} & 0.809 & 0.887 & 0.228 & \textbf{0.803} & 0.356 & 0.622 & 0.808 & 0.282 \\
 & & PsyCoT & 0.972 & \textbf{0.965} & \textbf{0.968} & \textbf{0.548} & 0.606 & \textbf{0.576} & \textbf{0.772} & \textbf{0.941} & \textbf{0.544} \\
\cmidrule(lr){2-12}
 & \multirow{3}{*}{DeepSeek-R1} & Vanilla & 0.984 & 0.849 & 0.912 & 0.273 & 0.803 & 0.408 & 0.660 & 0.846 & 0.343 \\
 & & CoT & \textbf{0.989} & 0.849 & 0.914 & 0.288 & \textbf{0.864} & 0.432 & 0.673 & 0.850 & 0.370 \\
 & & PsyCoT & 0.970 & \textbf{0.984} & \textbf{0.977} & \textbf{0.717} & 0.576 & \textbf{0.639} & \textbf{0.808} & \textbf{0.957} & \textbf{0.616} \\
\midrule

\multirow{12}{*}{Generalized Anxiety} 
 & \multirow{3}{*}{GPT-4o} & Vanilla & 0.953 & 0.996 & \textbf{0.974} & 0.979 & 0.777 & 0.866 & 0.920 & \textbf{0.957} & 0.841 \\
 & & CoT & \textbf{0.956} & 0.994 & \textbf{0.974} & 0.966 & \textbf{0.788} & \textbf{0.868} & \textbf{0.921} & \textbf{0.957} & \textbf{0.842} \\
 & & PsyCoT & 0.949 & \textbf{0.999} & 0.973 & \textbf{0.993} & 0.754 & 0.857 & 0.915 & 0.955 & 0.831 \\
\cmidrule(lr){2-12}
 & \multirow{3}{*}{Claude-3.5} & Vanilla & 0.953 & 0.987 & 0.970 & 0.927 & 0.777 & 0.845 & 0.907 & 0.949 & 0.815 \\
 & & CoT & 0.959 & 0.985 & 0.972 & 0.923 & 0.804 & 0.860 & 0.916 & 0.953 & 0.832 \\
 & & PsyCoT & \textbf{0.963} & \textbf{0.999} & \textbf{0.980} & \textbf{0.993} & \textbf{0.821} & \textbf{0.899} & \textbf{0.940} & \textbf{0.967} & \textbf{0.880} \\
\cmidrule(lr){2-12}
 & \multirow{3}{*}{GLM-Zero} & Vanilla & 0.984 & 0.883 & 0.931 & 0.635 & 0.933 & 0.756 & 0.843 & 0.892 & 0.690 \\
 & & CoT & \textbf{0.985} & 0.897 & 0.939 & 0.664 & \textbf{0.939} & 0.778 & 0.858 & 0.904 & 0.719 \\
 & & PsyCoT & 0.981 & \textbf{0.989} & \textbf{0.985} & \textbf{0.948} & 0.911 & \textbf{0.929} & \textbf{0.957} & \textbf{0.975} & \textbf{0.914} \\
\cmidrule(lr){2-12}
 & \multirow{3}{*}{DeepSeek-R1} & Vanilla & 0.961 & 0.978 & 0.969 & 0.890 & 0.816 & 0.851 & 0.910 & 0.949 & 0.821 \\
 & & CoT & 0.962 & 0.987 & 0.974 & 0.930 & 0.821 & 0.872 & 0.923 & 0.957 & 0.847 \\
 & & PsyCoT & \textbf{0.986} & \textbf{0.994} & \textbf{0.990} & \textbf{0.971} & \textbf{0.933} & \textbf{0.952} & \textbf{0.971} & \textbf{0.983} & \textbf{0.941} \\
\midrule

\multirow{12}{*}{Social Anxiety} 
 & \multirow{3}{*}{GPT-4o} & Vanilla & \textbf{0.999} & 0.969 & 0.984 & 0.610 & \textbf{0.979} & 0.752 & 0.868 & 0.969 & 0.736 \\
 & & CoT & 0.998 & 0.968 & 0.982 & 0.597 & 0.958 & 0.736 & 0.859 & 0.967 & 0.719 \\
 & & PsyCoT & 0.984 & \textbf{1.000} & \textbf{0.992} & \textbf{1.000} & 0.667 & \textbf{0.800} & \textbf{0.896} & \textbf{0.984} & \textbf{0.792} \\
\cmidrule(lr){2-12}
 & \multirow{3}{*}{Claude-3.5} & Vanilla & \textbf{1.000} & 0.950 & 0.974 & 0.500 & \textbf{1.000} & 0.667 & 0.820 & 0.952 & 0.644 \\
 & & CoT & \textbf{1.000} & 0.954 & 0.976 & 0.522 & \textbf{1.000} & 0.686 & 0.831 & 0.956 & 0.665 \\
 & & PsyCoT & 0.984 & \textbf{0.997} & \textbf{0.991} & \textbf{0.917} & 0.688 & \textbf{0.786} & \textbf{0.888} & \textbf{0.982} & \textbf{0.777} \\
\cmidrule(lr){2-12}
 & \multirow{3}{*}{GLM-Zero} & Vanilla & \textbf{1.000} & 0.901 & 0.948 & 0.338 & \textbf{1.000} & 0.505 & 0.727 & 0.906 & 0.467 \\
 & & CoT & \textbf{1.000} & 0.907 & 0.951 & 0.350 & \textbf{1.000} & 0.519 & 0.735 & 0.911 & 0.482 \\
 & & PsyCoT & 0.996 & \textbf{0.994} & \textbf{0.995} & \textbf{0.880} & 0.917 & \textbf{0.898} & \textbf{0.946} & \textbf{0.990} & \textbf{0.893} \\
\cmidrule(lr){2-12}
 & \multirow{3}{*}{DeepSeek-R1} & Vanilla & \textbf{1.000} & 0.950 & 0.974 & 0.500 & \textbf{1.000} & 0.667 & 0.820 & 0.952 & 0.644 \\
 & & CoT & \textbf{1.000} & 0.950 & 0.974 & 0.500 & \textbf{1.000} & 0.667 & 0.820 & 0.952 & 0.644 \\
 & & PsyCoT & {0.997} & \textbf{0.995} & \textbf{0.996} & \textbf{0.900} & 0.938 & \textbf{0.918} & \textbf{0.957} & \textbf{0.992} & \textbf{0.914} \\
\midrule

\multirow{12}{*}{Suicide Risk} 
 & \multirow{3}{*}{GPT-4o} & Vanilla & 0.933 & \textbf{0.999} & 0.965 & \textbf{0.955} & 0.241 & 0.385 & 0.675 & 0.933 & 0.363 \\
 & & CoT & 0.930 & \textbf{0.999} & 0.963 & 0.947 & 0.207 & 0.340 & 0.651 & 0.930 & 0.318 \\
 & & PsyCoT & \textbf{0.993} & 0.996 & \textbf{0.995} & 0.953 & \textbf{0.931} & \textbf{0.942} & \textbf{0.968} & \textbf{0.990} & \textbf{0.936} \\
\cmidrule(lr){2-12}
 & \multirow{3}{*}{Claude-3.5} & Vanilla & 0.937 & 0.991 & 0.963 & 0.765 & 0.299 & 0.430 & 0.697 & 0.931 & 0.400 \\
 & & CoT & 0.938 & 0.995 & 0.966 & 0.844 & 0.310 & 0.454 & 0.710 & 0.935 & 0.427 \\
 & & PsyCoT & \textbf{0.978} & \textbf{0.998} & \textbf{0.988} & \textbf{0.971} & \textbf{0.759} & \textbf{0.852} & \textbf{0.920} & \textbf{0.977} & \textbf{0.839} \\
\cmidrule(lr){2-12}
 & \multirow{3}{*}{GLM-Zero} & Vanilla & 0.933 & 0.950 & 0.941 & 0.352 & 0.287 & 0.316 & 0.629 & 0.892 & 0.259 \\
 & & CoT & 0.936 & 0.938 & 0.937 & 0.329 & 0.322 & 0.326 & 0.631 & 0.884 & 0.262 \\
 & & PsyCoT & \textbf{0.997} & \textbf{0.992} & \textbf{0.995} & \textbf{0.923} & \textbf{0.966} & \textbf{0.944} & \textbf{0.969} & \textbf{0.990} & \textbf{0.938} \\
\cmidrule(lr){2-12}
 & \multirow{3}{*}{DeepSeek-R1} & Vanilla & 0.934 & \textbf{0.998} & 0.965 & 0.917 & 0.253 & 0.396 & 0.681 & 0.933 & 0.373 \\
 & & CoT & {0.935} & \textbf{0.998} & 0.966 & {0.923} & 0.276 & 0.425 & {0.695} & {0.935} & {0.401} \\
 & & PsyCoT & \textbf{0.992} & \textbf{0.998} & \textbf{0.995} & \textbf{0.976} & \textbf{0.920} & \textbf{0.947} & \textbf{0.971} & \textbf{0.991} & \textbf{0.942} \\
\bottomrule
\end{tabular}}
\caption{Comparative performance analysis of LLMs with zero-shot prompting strategies.}
\label{tab:zero-shot-main_results}
\end{table*}

%% file: Appendix/fewshot_results.tex
\subsection{Few-shot Results}
\label{sec:app-fewshot}
The few-shot analysis demonstrates enhanced diagnostic capabilities through in-context learning, with PsyCoT achieving state-of-the-art performance across most metrics as shown in Table \ref{tab:app_fewshot_results}. For suicide risk assessment, a critical clinical task, PsyCoT improved GPT-4o's case F1 from 0.500 to 0.927, an 85\% increase, and Cohen's $\kappa$ from 0.475 to 0.920, a 93\% improvement compared to vanilla prompting. This approach achieved near-perfect control group recall of 0.990 while maintaining high precision of 0.902. Model specialization emerged in the results: Claude-3.5 attained peak anxiety detection performance with a Macro-F1 of 0.934 using 3-shot PsyCoT, while GLM-Zero showed dramatic improvements in depression classification with Macro-F1 increasing by 19\% to 0.788 and $\kappa$ increasing by 72\% to 0.580.
DeepSeek-R1 achieved the highest social anxiety detection scores with a Macro-F1 of 0.973 through vanilla few-shot learning, suggesting certain architectures benefit more from direct exemplar exposure. However, PsyCoT universally enhanced inter-rater reliability, particularly for suicide risk where $\kappa$ values exceeded 0.900 across all models—a threshold indicating clinical-grade consensus.
The results reveal an inverse precision-recall tradeoff: while control group precision remained stable at 0.987 with a standard deviation of 0.011 across disorders, case recall showed greater variance, ranging from 0.485 to 0.954 for depression in PsyCoT. This highlights the need for task-specific calibration when deploying few-shot models in clinical practice.
\begin{table*}[htbp]
\centering
\resizebox{\textwidth}{!}{
\begin{tabular}{llllllllllll}
\toprule
\multirow{2}{*}{Disorder} & \multirow{2}{*}{Model} & \multirow{2}{*}{Method} & \multicolumn{3}{c}{Control} & \multicolumn{3}{c}{Case} & \multirow{2}{*}{Macro-F1} & \multirow{2}{*}{Acc.} & \multirow{2}{*}{$\kappa$} \\
\cmidrule(lr){4-6} \cmidrule(lr){7-9}
 & & & P & R & F1 & P & R & F1 & & & \\
\midrule
\multirow{12}{*}{Depression} 
 & \multirow{3}{*}{GPT-4o} & Vanilla & 0.961 & 0.951 & 0.956 & 0.395 & 0.455 & 0.423 & 0.689 & 0.918 & 0.379 \\
 & & CoT & 0.962 & 0.957 & 0.960 & 0.437 & 0.470 & 0.453 & 0.706 & 0.925 & 0.412 \\
 & & PsyCoT & \textbf{0.965} & \textbf{0.991} & \textbf{0.978} & \textbf{0.800} & \textbf{0.485} & \textbf{0.604} & \textbf{0.791} & \textbf{0.958} & \textbf{0.583} \\
\cmidrule(lr){2-12}
 & \multirow{3}{*}{Claude-3.5} & Vanilla & \textbf{0.971} & 0.933 & 0.951 & 0.388 & \textbf{0.606} & 0.473 & 0.712 & 0.911 & 0.427 \\
 & & CoT & {0.970} & 0.948 & 0.959 & 0.443 & 0.591 & 0.506 & 0.733 & 0.924 & 0.466 \\
 & & PsyCoT & {0.967} & \textbf{0.994} & \textbf{0.980} & \textbf{0.850} & {0.515} & \textbf{0.642} & \textbf{0.811} & \textbf{0.962} & \textbf{0.623} \\
\cmidrule(lr){2-12}
 & \multirow{3}{*}{GLM-Zero} & Vanilla & 0.974 & 0.881 & 0.925 & 0.284 & 0.667 & 0.398 & 0.662 & 0.867 & 0.337 \\
 & & CoT & 0.975 & 0.878 & 0.924 & 0.283 & 0.682 & 0.400 & 0.662 & 0.865 & 0.338 \\
 & & PsyCoT & \textbf{0.988} & \textbf{0.938} & \textbf{0.962} & \textbf{0.487} & \textbf{0.833} & \textbf{0.615} & \textbf{0.788} & \textbf{0.931} & \textbf{0.580} \\
\cmidrule(lr){2-12}
 & \multirow{3}{*}{DeepSeek-R1} & Vanilla & 0.966 & 0.978 & 0.972 & 0.618 & 0.515 & 0.562 & 0.767 & 0.947 & 0.534 \\
 & & CoT & \textbf{0.969} & 0.973 & 0.971 & 0.597 & \textbf{0.561} & 0.578 & 0.775 & 0.946 & 0.549 \\
 & & PsyCoT & \textbf{0.969} & \textbf{0.989} & \textbf{0.979} & \textbf{0.783} & {0.545} & \textbf{0.643} & \textbf{0.811} & \textbf{0.960} & \textbf{0.622} \\
\midrule

\multirow{12}{*}{Generalized Anxiety} 
 & \multirow{3}{*}{GPT-4o} & Vanilla & 0.955 & 0.998 & 0.976 & 0.986 & 0.782 & 0.872 & 0.924 & 0.959 & 0.848 \\
 & & CoT & 0.956 & 0.996 & 0.976 & 0.979 & 0.788 & 0.873 & 0.924 & 0.959 & 0.849 \\
 & & PsyCoT & \textbf{0.975} & \textbf{1.000} & \textbf{0.987} & \textbf{1.000} & \textbf{0.883} & \textbf{0.938} & \textbf{0.963} & \textbf{0.979} & \textbf{0.925} \\
\cmidrule(lr){2-12}
 & \multirow{3}{*}{Claude-3.5} & Vanilla & \textbf{0.973} & 0.978 & 0.976 & 0.897 & \textbf{0.877} & 0.887 & 0.931 & 0.960 & 0.863 \\
 & & CoT & 0.971 & 0.977 & 0.974 & 0.891 & 0.866 & 0.878 & 0.926 & 0.957 & 0.852 \\
 & & PsyCoT & {0.959} & \textbf{0.999} & \textbf{0.979} & \textbf{0.993} & {0.804} & \textbf{0.889} & \textbf{0.934} & \textbf{0.964} & \textbf{0.868} \\
\cmidrule(lr){2-12}
 & \multirow{3}{*}{GLM-Zero} & Vanilla & 0.986 & 0.944 & 0.965 & 0.785 & 0.939 & 0.855 & 0.910 & 0.943 & 0.820 \\
 & & CoT & 0.987 & 0.948 & 0.967 & 0.797 & 0.944 & 0.864 & 0.916 & 0.947 & 0.832 \\
 & & PsyCoT & \textbf{0.991} & \textbf{0.984} & \textbf{0.988} & \textbf{0.930} & \textbf{0.961} & \textbf{0.945} & \textbf{0.966} & \textbf{0.980} & \textbf{0.933} \\
\cmidrule(lr){2-12}
 & \multirow{3}{*}{DeepSeek-R1} & Vanilla & 0.966 & 0.973 & 0.970 & 0.873 & 0.844 & 0.858 & 0.914 & 0.950 & 0.828 \\
 & & CoT & 0.961 & 0.978 & 0.969 & 0.890 & 0.816 & 0.851 & 0.910 & 0.949 & 0.821 \\
 & & PsyCoT & \textbf{0.989} & \textbf{0.994} & \textbf{0.992} & \textbf{0.971} & \textbf{0.950} & \textbf{0.960} & \textbf{0.976} & \textbf{0.986} & \textbf{0.952} \\
\midrule

\multirow{12}{*}{Social Anxiety} 
 & \multirow{3}{*}{GPT-4o} & Vanilla & \textbf{0.993} & \textbf{1.000} & \textbf{0.996} & \textbf{1.000} & \textbf{0.854} & \textbf{0.921} & \textbf{0.959} & \textbf{0.993} & \textbf{0.918} \\
 & & CoT & 0.992 & \textbf{1.000} & \textbf{0.996} & \textbf{1.000} & 0.833 & 0.909 & 0.952 & 0.992 & 0.905 \\
 & & PsyCoT & 0.988 & \textbf{1.000} & 0.994 & \textbf{1.000} & {0.750} & {0.857} & {0.925} & {0.988} & {0.851} \\
\cmidrule(lr){2-12}
 & \multirow{3}{*}{Claude-3.5} & Vanilla & 0.991 & 0.993 & 0.992 & 0.848 & 0.813 & 0.830 & 0.911 & 0.984 & 0.821 \\
 & & CoT & \textbf{0.992} & {0.992} & 0.992 & {0.833} & \textbf{0.833} & \textbf{0.833} & 0.912 & 0.984 & 0.825 \\
 & & PsyCoT & {0.987} & \textbf{0.999} & \textbf{0.993} & \textbf{0.972} & {0.729} & \textbf{0.833} & \textbf{0.913} & \textbf{0.986} & \textbf{0.826} \\
\cmidrule(lr){2-12}
 & \multirow{3}{*}{GLM-Zero} & Vanilla & \textbf{1.000} & 0.958 & 0.979 & 0.545 & \textbf{1.000} & 0.706 & 0.842 & 0.960 & 0.686 \\
 & & CoT & \textbf{1.000} & 0.965 & 0.982 & 0.593 & \textbf{1.000} & 0.744 & 0.863 & 0.967 & 0.728 \\
 & & PsyCoT & {0.995} & \textbf{0.996} & \textbf{0.995} & \textbf{0.915} & {0.896} & \textbf{0.905} & \textbf{0.950} & \textbf{0.991} & \textbf{0.901} \\
\cmidrule(lr){2-12}
 & \multirow{3}{*}{DeepSeek-R1} & Vanilla & \textbf{0.999} & {0.996} & {0.997} & {0.922} & \textbf{0.979} & {0.949} & {0.973} & {0.995} & {0.947} \\
 & & CoT & 0.997 & 0.994 & 0.995 & 0.882 & 0.938 & 0.909 & 0.952 & 0.991 & 0.904 \\
 & & PsyCoT & 0.998 & \textbf{0.999} & \textbf{0.998} & \textbf{0.979} & 0.958 & \textbf{0.968} & \textbf{0.983} & \textbf{0.997} & \textbf{0.967} \\
\midrule

\multirow{12}{*}{Suicide Risk} 
 & \multirow{3}{*}{GPT-4o} & Vanilla & 0.941 & \textbf{0.997} & 0.968 & \textbf{0.909} & 0.345 & 0.500 & 0.734 & 0.940 & 0.475 \\
 & & CoT & 0.940 & \textbf{0.997} & 0.968 & 0.906 & 0.333 & 0.487 & 0.728 & 0.939 & 0.462 \\
 & & PsyCoT & \textbf{0.996} & {0.990} & \textbf{0.993} & {0.902} & \textbf{0.954} & \textbf{0.927} & \textbf{0.960} & \textbf{0.987} & \textbf{0.920} \\
\cmidrule(lr){2-12}
 & \multirow{3}{*}{Claude-3.5} & Vanilla & 0.960 & 0.960 & 0.960 & 0.575 & 0.575 & 0.575 & 0.767 & 0.926 & 0.534 \\
 & & CoT & 0.948 & 0.958 & 0.953 & 0.506 & 0.448 & 0.476 & 0.714 & 0.914 & 0.429 \\
 & & PsyCoT & \textbf{0.990} & \textbf{0.993} & \textbf{0.992} & \textbf{0.929} & \textbf{0.897} & \textbf{0.912} & \textbf{0.952} & \textbf{0.985} & \textbf{0.904} \\
\cmidrule(lr){2-12}
 & \multirow{3}{*}{GLM-Zero} & Vanilla & 0.938 & 0.983 & 0.960 & 0.636 & 0.322 & 0.427 & 0.694 & 0.925 & 0.392 \\
 & & CoT & 0.938 & 0.984 & 0.960 & 0.643 & 0.310 & 0.419 & 0.689 & 0.925 & 0.384 \\
 & & PsyCoT & \textbf{0.997} & \textbf{0.993} & \textbf{0.995} & \textbf{0.933} & \textbf{0.966} & \textbf{0.949} & \textbf{0.972} & \textbf{0.991} & \textbf{0.944} \\
\cmidrule(lr){2-12}
 & \multirow{3}{*}{DeepSeek-R1} & Vanilla & 0.950 & 0.995 & 0.972 & 0.886 & 0.448 & 0.595 & 0.784 & 0.947 & 0.570 \\
 & & CoT & 0.944 & 0.990 & 0.966 & 0.786 & 0.379 & 0.512 & 0.739 & 0.937 & 0.482 \\
 & & PsyCoT & \textbf{0.991} & \textbf{0.998} & \textbf{0.995} & \textbf{0.975} & \textbf{0.908} & \textbf{0.940} & \textbf{0.968} & \textbf{0.990} & \textbf{0.935} \\
\bottomrule
\end{tabular}}
\caption{Comparative performance analysis of LLMs with few-shot prompting strategies.}
\label{tab:app_fewshot_results}
\end{table*}

%% file: Appendix/demo.tex
\section{Demo}
\label{app:demo}

\begin{figure*}[ht]
    \centering
    \includegraphics[width=1\linewidth]{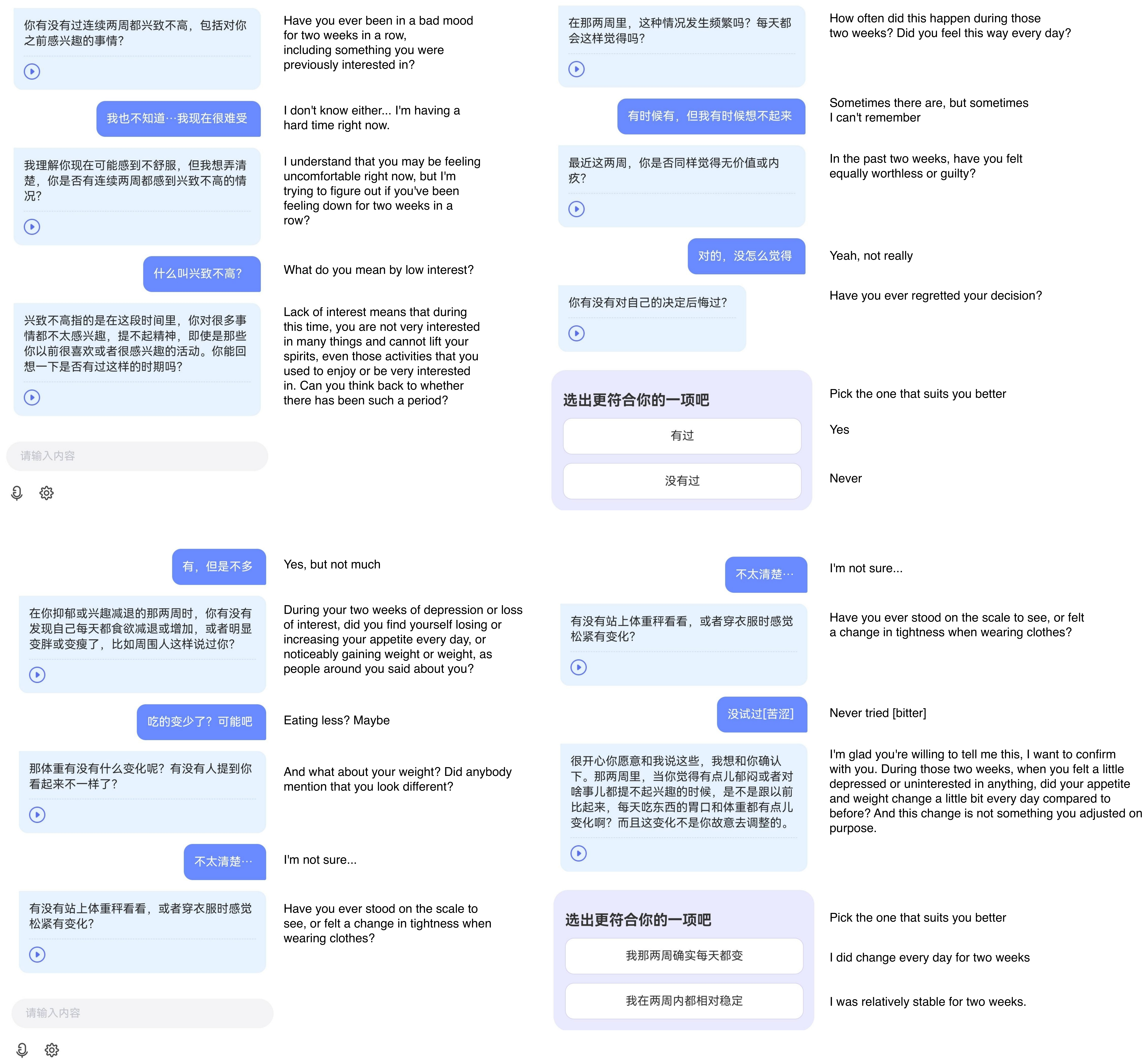}
    \caption{Case of the demo of MAGI.}
    \label{fig:demo}
\end{figure*}

We introduce the demo of \method, which has been successfully deployed and widely adopted by a large user base. The chatbot offers questions in both text and audio formats, ensuring an inclusive experience for users with diverse needs. Through this interactive and intuitive interface, our system supports self-reflection and early mental health screening, positioning it as a valuable tool for large-scale implementation.

Real-world usage has shown that our system not only helps users articulate their emotions effectively but also aligns seamlessly with established diagnostic frameworks, demonstrating its potential to make a meaningful impact on mental health awareness and care.

%% file: Appendix/scale.tex
\onecolumn
\section{Scale Guidebook}
\label{app:scale}
\begin{tcolorbox}[title=AI Psychological Interview Dialogue Validity Rating Scale (1-5 points),enhanced,breakable]

\textbf{Relevance}
\begin{itemize}[noitemsep, topsep=0pt]
    \item 5 : AI responses are consistently closely related to user questions, emotions, and topics, highly aligned with the psychological interview context, and accurately address core user content.
    \item 4 : Most responses are closely related to user expressions, with only a few minor deviations that do not affect the overall communication.
    \item 3 : Some responses are related to user topics, but there are some deviations, although basic communication can still be maintained.
    \item 2 : Many responses deviate from user topics, with insufficient understanding of core user needs, and obvious obstacles in communication.
    \item 1 point: AI responses are almost completely detached from user topics, seriously disconnected from the psychological interview context, and frequently miss the point.
\end{itemize}

\textbf{Accuracy}
\begin{itemize}[noitemsep, topsep=0pt]
    \item 5 : Precise understanding of user questions, suggestions, and analysis provided are fully in line with psychological principles and common sense, accurately grasp and respond to user emotions.
    \item 4 : Basically accurately understand user questions, with slight deviations in the application of psychological knowledge and emotional judgment, but does not affect overall understanding.
    \item 3 : Able to understand some user questions, but there are more deviations in understanding, and the suggestions given are somewhat reasonable but not rigorous enough.
    \item 2 : Frequently misunderstand user questions, the information provided has many errors and may mislead users, and the emotional judgment is inaccurate.
    \item 1 point: Frequently and seriously misunderstand user questions, the information given completely violates psychological principles, seriously misleads users, and is unable to recognize emotions.
\end{itemize}

\textbf{Completeness}
\begin{itemize}[noitemsep, topsep=0pt]
    \item 5 : The response content is complete, well-organized and clearly structured, comprehensively covering all aspects of the user's question, with no omission of key information.
    \item 4 : The response content is relatively complete, including most of the key information, with slight deficiencies in details.
    \item 3 : Able to provide some key information, but there are many deficiencies in the overall content, and the logical structure is not clear enough.
    \item 2 : The response content is severely missing key information, too brief and vague, making it difficult for users to obtain effective help.
    \item 1 point: The response content is chaotic and provides almost no valuable information, completely failing to meet user needs.
\end{itemize}

\textbf{Guidance}
\begin{itemize}[noitemsep, topsep=0pt]
    \item 5 : Actively and proactively guides users to express themselves in depth, effectively promotes the in-depth development of the interview through appropriate methods, and has strong interactivity.
    \item 4 : Able to guide, the guidance method is basically natural, and it has a certain role in promoting users' in-depth expression.
    \item 3 : Occasionally guide users, the guidance effect is average, and sometimes it is not possible to effectively stimulate users to express themselves further.
    \item 2 : Rarely actively guide users, the interview is more passive, lacking interactivity and exploration.
    \item 1 point: Completely lacks the intention to guide users, only mechanically responds to user questions, and the interview is dull and uninteresting.
\end{itemize}

\end{tcolorbox}

\twocolumn

%% file: Appendix/prompt.tex
\clearpage
\onecolumn
\section{Prompts}
% \subsection{Prompt for Depression}
\begin{tcolorbox}[title=Depression Detection Prompt,enhanced,breakable]

\section*{Input}
\begin{itemize}[noitemsep, topsep=0pt]
    \item Provide psychological counseling dialogue content.
    \item Determine whether the dialogue content meets the following depression diagnostic criteria and strictly follow the reasoning process.
\end{itemize}

\section*{Symptom Definitions}

\subsection*{1. Depressed Mood}
\textbf{Symptom Manifestation:} Subjectively feeling sad, empty, hopeless, or experiencing emotional distress; or observed crying, emotional breakdowns, or shouting.

\textbf{Corresponding Questions:}
\begin{itemize}[noitemsep, topsep=0pt]
    \item "In your life, have you ever had a period of at least two consecutive weeks where you couldn't feel happy, felt sad, empty, or hopeless every day?"
    \item "That period must have been tough for you. Can you tell me how that felt during those two weeks? Was there anything specific that triggered these feelings?"
    \item "In the past two weeks, have you felt down or empty every day?"
\end{itemize}

\textbf{Scope of This Symptom:}
\begin{itemize}[noitemsep, topsep=0pt]
\item \textbf{Time Range:} Must last for at least two weeks.
\item \textbf{Time Frequency:} Must occur "most of the day, nearly every day."
\item \textbf{Symptom Reasoning Process:}
\begin{itemize}[noitemsep, topsep=0pt]
\item Both question 1 and question 3 must be answered affirmatively.
\item If question 2 is present, the described feelings must align with the symptom manifestations.
\item Only when these conditions are met, can the symptom be judged as "Yes."
\end{itemize}
\end{itemize}

\subsection*{2. Loss of Interest or Pleasure}
\textbf{Symptom Manifestation:} Significant decrease in interest or pleasure in almost all activities.

\textbf{Corresponding Questions:}
\begin{itemize}[noitemsep, topsep=0pt]
    \item "Have you ever had a period of at least two consecutive weeks where you had little interest, even in things you used to enjoy?"
    \item "In the past two weeks, have you also been feeling uninterested and finding things meaningless?"
    \item "When you feel uninterested, how do you usually spend your time?"
\end{itemize}

\textbf{Scope of This Symptom:}
\begin{itemize}[noitemsep, topsep=0pt]
    \item \textbf{Time Range:} Must last for at least two weeks.
    \item \textbf{Time Frequency:} Must occur "most of the day, nearly every day."
    \item \textbf{Symptom Reasoning Process:}
        \begin{itemize}[noitemsep, topsep=0pt]
            \item Both question 1 and question 2 must be answered affirmatively.
            \item If question 3 is present, the response should confirm a lack of interest, especially in previously enjoyable activities.
            \item Only when these conditions are met, can the symptom be judged as "Yes."
        \end{itemize}
\end{itemize}

\subsection*{3. Weight or Appetite Change}
\textbf{Symptom Manifestation:} Significant weight loss or gain, or decreased/increased appetite, without intentional dieting.

\textbf{Corresponding Questions:}
\begin{itemize}[noitemsep, topsep=0pt]
    \item "During the two weeks of feeling depressed or uninterested, did you notice daily appetite changes or significant weight fluctuations? Did people around you comment on this?"
    \item "In the past two weeks, has your eating or weight changed?"
    \item "Have your eating habits and weight remained stable in the past two weeks? Any noticeable differences?"
    \item "Compared to before, has your food intake changed significantly in the past two weeks?"
    \item "When you felt down or uninterested, did you experience such changes daily in eating habits or weight, even though you were not intentionally dieting or overeating?"
\end{itemize}

\textbf{Scope of This Symptom:}
\begin{itemize}[noitemsep, topsep=0pt]
    \item \textbf{Time Range:} Within a past depressive episode, not limited to the recent period.
    \item \textbf{Symptom Reasoning Process:}
        \begin{itemize}[noitemsep, topsep=0pt]
            \item Both question 1 and question 5 must be answered affirmatively.
            \item Questions 2, 3, and 4 should indicate unstable weight, significant appetite changes, or altered eating habits.
            \item Only when these conditions are met, can the symptom be judged as "Yes."
        \end{itemize}
\end{itemize}

\subsection*{4. Sleep Disturbance}
\textbf{Symptom Manifestation:} Insomnia or excessive sleep.

\textbf{Corresponding Questions:}
\begin{itemize}[noitemsep, topsep=0pt]
    \item "During the two weeks when you felt depressed or lost interest, did you have trouble sleeping every night, such as difficulty falling asleep, waking up in the middle of the night, waking up too early, or sleeping too much?"
    \item "In the past two weeks, have you experienced the same sleep problems?"
\end{itemize}

\textbf{Scope of This Symptom:}
\begin{itemize}[noitemsep, topsep=0pt]
    \item \textbf{Time Range:} Within a past depressive episode.
    \item \textbf{Time Frequency:} "Sometimes" or more frequently.
    \item \textbf{Corresponding Question Range:}
        \begin{itemize}[noitemsep, topsep=0pt]
            \item "During those two weeks of feeling depressed or uninterested, did you have trouble sleeping every night, such as difficulty falling asleep, waking up in the middle of the night, waking too early, or sleeping too much?"
            \item "In the past two weeks, have you had the same sleep problems?"
        \end{itemize}
\end{itemize}

\subsection*{5. Agitation or Sluggishness in Daily Behavior}
\textbf{Symptom Manifestation:} Physical agitation or sluggishness, or feeling mentally dull or numb.

\textbf{Corresponding Questions:}
\begin{itemize}[noitemsep, topsep=0pt]
    \item "During the two weeks of feeling depressed or uninterested, did you become sluggish or agitated almost every day? Were you noticeably restless or unable to sit still?"
    \item "In the past two weeks, have you had similar feelings of sluggishness or agitation?"
\end{itemize}

\textbf{Scope of This Symptom:}
\begin{itemize}[noitemsep, topsep=0pt]
    \item \textbf{Time Range:} Within a past depressive episode.
    \item \textbf{Time Frequency:} "Sometimes" or more frequently.
    \item \textbf{Symptom Reasoning Process:}
        \begin{itemize}[noitemsep, topsep=0pt]
            \item Both questions 1 and 2 must be answered affirmatively.
            \item This condition would indicate either physical restlessness or emotional numbness.
            \item Only when these conditions are met, can the symptom be judged as "Yes."
        \end{itemize}
\end{itemize}

\subsection*{6. Fatigue or Loss of Energy}
\textbf{Symptom Manifestation:} Persistent fatigue or a lack of energy, feeling drained.

\textbf{Corresponding Questions:}
\begin{itemize}[noitemsep, topsep=0pt]
    \item "During the two weeks of feeling depressed or uninterested, did you feel like you had no energy, almost every day?"
    \item "In the past two weeks, have you had similar feelings of being constantly tired or lacking energy?"
    \item "Was the lack of energy constant or did it occur sporadically?"
\end{itemize}

\textbf{Scope of This Symptom:}
\begin{itemize}[noitemsep, topsep=0pt]
    \item \textbf{Time Range:} Within a past depressive episode.
    \item \textbf{Time Frequency:} "Sometimes" or more frequently.
    \item \textbf{Symptom Reasoning Process:}
        \begin{itemize}[noitemsep, topsep=0pt]
            \item If both questions are answered affirmatively, it suggests significant lack of energy.
            \item The symptoms must be noticeable almost every day.
            \item Only when these conditions are met, can the symptom be judged as "Yes."
        \end{itemize}
\end{itemize}

\subsection*{7. Feelings of Worthlessness or Excessive Guilt}
\textbf{Symptom Manifestation:} Feeling worthless or excessively guilty, often inappropriate guilt.

\textbf{Corresponding Questions:}
\begin{itemize}[noitemsep, topsep=0pt]
    \item "During the two weeks of feeling depressed or uninterested, did you feel almost every day that you were worthless or excessively guilty?"
    \item "In the past two weeks, have you felt the same sense of worthlessness or guilt?"
\end{itemize}

\textbf{Scope of This Symptom:}
\begin{itemize}[noitemsep, topsep=0pt]
    \item \textbf{Time Range:} Within a past depressive episode.
    \item \textbf{Time Frequency:} "Sometimes" or more frequently.
    \item \textbf{Symptom Reasoning Process:}
        \begin{itemize}[noitemsep, topsep=0pt]
            \item Both questions must be answered affirmatively to indicate significant feelings of worthlessness or guilt.
            \item Only when these conditions are met, can the symptom be judged as "Yes."
        \end{itemize}
\end{itemize}

\subsection*{8. Difficulty Concentrating or Making Decisions}
\textbf{Symptom Manifestation:} Difficulty thinking clearly or making decisions, inability to focus.

\textbf{Corresponding Questions:}
\begin{itemize}[noitemsep, topsep=0pt]
    \item "During the two weeks of feeling depressed or uninterested, did you often find it difficult to concentrate or make decisions?"
    \item "In the past two weeks, have you had trouble thinking clearly or making decisions?"
\end{itemize}

\textbf{Scope of This Symptom:}
\begin{itemize}[noitemsep, topsep=0pt]
    \item \textbf{Time Range:} Within a past depressive episode.
    \item \textbf{Time Frequency:} "Sometimes" or more frequently.
    \item \textbf{Symptom Reasoning Process:}
        \begin{itemize}[noitemsep, topsep=0pt]
            \item If both questions are answered affirmatively, it indicates difficulty concentrating or decision-making.
            \item Only when these conditions are met, can the symptom be judged as "Yes."
        \end{itemize}
\end{itemize}

\subsection*{9. Thoughts of Death or Suicide}
\textbf{Symptom Manifestation:} Frequent thoughts of death, suicidal ideation, or actual suicide attempts.

\textbf{Corresponding Questions:}
\begin{itemize}[noitemsep, topsep=0pt]
    \item "During the two weeks of feeling depressed or uninterested, did you often think about death or have thoughts of suicide, or even plan or attempt suicide?"
    \item "In the past two weeks, have you had thoughts of death, suicide, or any suicide plans or attempts?"
\end{itemize}

\textbf{Scope of This Symptom:}
\begin{itemize}[noitemsep, topsep=0pt]
    \item \textbf{Time Range:} Within a past depressive episode.
    \item \textbf{Time Frequency:} At least once.
    \item \textbf{Symptom Reasoning Process:}
        \begin{itemize}[noitemsep, topsep=0pt]
            \item If either question is answered negatively, it would indicate the absence of thoughts of death or suicide.
            \item If the answer is affirmative for either question, the symptom can be judged as "Yes."
        \end{itemize}
\end{itemize}

\subsection*{10. Clinically Significant Distress or Impairment}
\textbf{Symptom Manifestation:} The symptoms cause significant distress or functional impairment.

\textbf{Corresponding Questions:}
\begin{itemize}[noitemsep, topsep=0pt]
    \item "During those two weeks, did these symptoms cause significant distress or problems in your life, such as at home, work, school, social activities, or relationships?"
    \item "In the past two weeks, have these symptoms also caused distress or problems in your life?"
\end{itemize}

\textbf{Scope of This Symptom:}
\begin{itemize}[noitemsep, topsep=0pt]
    \item \textbf{Symptom Reasoning Process:} A score-based system determines if the impairment is significant.
\end{itemize}

\subsection*{\textcolor{red}{Diagnosis Criteria}}
\begin{itemize}[noitemsep, topsep=0pt]
    \item \textbf{Symptoms 1-9:} At least \textbf{five} symptoms must be present, including either \textbf{depressed mood or loss of interest}.
    \item \textbf{Symptom 10:} Must be present.
\end{itemize}
\end{tcolorbox}

\begin{tcolorbox}[title=Generalized Anxiety Disorder Detection Prompt,enhanced,breakable]

\subsection*{1. General Anxiety Across Multiple Issues}
\textbf{Symptom Manifestation:} The individual expresses anxiety about multiple issues.

\textbf{Corresponding Questions:}
\begin{itemize}[noitemsep, topsep=0pt]
    \item "Do you feel anxious about several things in your life?"
\end{itemize}

\textbf{Scope of This Symptom:}
\begin{itemize}[noitemsep, topsep=0pt]
    \item \textbf{Symptom Reasoning Process:} If the user mentions anxiety across multiple areas, mark as "Yes". If they only mention one specific concern and the response is affirmative, mark as "Yes". If no anxiety is expressed, mark as "No".
    \item If the answer is unclear, mark as "Uncertain".
\end{itemize}

\subsection*{2. Anxiety in the Past 6 Months}
\textbf{Symptom Manifestation:} The individual has experienced excessive anxiety for more than half of the days over the past six months.

\textbf{Corresponding Questions:}
\begin{itemize}[noitemsep, topsep=0pt]
    \item "Have you been anxious for most of the past six months regarding work, school, or other activities?"
\end{itemize}

\textbf{Scope of This Symptom:}
\begin{itemize}[noitemsep, topsep=0pt]
    \item \textbf{Symptom Reasoning Process:} If the user expresses significant anxiety for a period of time in the past six months, mark as "Yes". If the anxiety is recent but severe, mark as "Yes". If less than six months of anxiety is reported and mild, mark as "No".
    \item If the answer is unclear, mark as "Uncertain".
\end{itemize}

\subsection*{3. Anxiety Frequency in the Past 6 Months}
\textbf{Symptom Manifestation:} The individual experiences frequent anxiety during the past six months, not just occasionally.

\textbf{Corresponding Questions:}
\begin{itemize}[noitemsep, topsep=0pt]
    \item "Did you experience significant anxiety more than just occasionally during the past few months?"
\end{itemize}

\textbf{Scope of This Symptom:}
\begin{itemize}[noitemsep, topsep=0pt]
    \item \textbf{Symptom Reasoning Process:} If the user reports frequent anxiety over a period of time in the last six months, mark as "Yes". If the anxiety is reported as occasional or only a few times, mark as "No".
    \item If the answer is unclear, mark as "Uncertain".
\end{itemize}

\subsection*{4. Difficulty Controlling Worry}
\textbf{Symptom Manifestation:} The individual finds it difficult to control their worry or anxiety.

\textbf{Corresponding Questions:}
\begin{itemize}[noitemsep, topsep=0pt]
    \item "Do you feel like you can't control your worries or anxiety?"
\end{itemize}

\textbf{Scope of This Symptom:}
\begin{itemize}[noitemsep, topsep=0pt]
    \item \textbf{Symptom Reasoning Process:} If the user states that they have difficulty controlling their anxiety or worry, mark as "Yes".
    \item If the answer is unclear or ambiguous, mark as "Uncertain".
\end{itemize}

\subsection*{5. Restlessness or Feeling on Edge}
\textbf{Symptom Manifestation:} The individual experiences restlessness, agitation, or tension.

\textbf{Corresponding Questions:}
\begin{itemize}[noitemsep, topsep=0pt]
    \item "Do you often feel restless, on edge, or agitated?"
\end{itemize}

\textbf{Scope of This Symptom:}
\begin{itemize}[noitemsep, topsep=0pt]
    \item \textbf{Symptom Reasoning Process:} If the user reports restlessness or a sense of being on edge, mark as "Yes".
    \item If the answer is unrelated to this symptom, mark as "Uncertain".
\end{itemize}

\subsection*{6. Muscle Tension}
\textbf{Symptom Manifestation:} The individual experiences muscle tension.

\textbf{Corresponding Questions:}
\begin{itemize}[noitemsep, topsep=0pt]
    \item "Do you experience muscle tension, such as tightness in your muscles?"
\end{itemize}

\textbf{Scope of This Symptom:}
\begin{itemize}[noitemsep, topsep=0pt]
    \item \textbf{Symptom Reasoning Process:} If the user confirms muscle tension, mark as "Yes".
    \item If the answer is unclear or vague, mark as "Uncertain".
\end{itemize}

\subsection*{7. Fatigue or Exhaustion}
\textbf{Symptom Manifestation:} The individual feels unusually tired or fatigued.

\textbf{Corresponding Questions:}
\begin{itemize}[noitemsep, topsep=0pt]
    \item "Do you feel easily fatigued or exhausted?"
\end{itemize}

\textbf{Scope of This Symptom:}
\begin{itemize}[noitemsep, topsep=0pt]
    \item \textbf{Symptom Reasoning Process:} If the user expresses fatigue, mark as "Yes". If the user states that it happens occasionally, mark as "Uncertain".
\end{itemize}

\subsection*{8. Difficulty Concentrating}
\textbf{Symptom Manifestation:} The individual has trouble concentrating or experiences a blank mind.

\textbf{Corresponding Questions:}
\begin{itemize}[noitemsep, topsep=0pt]
    \item "Do you find it hard to concentrate or focus, or feel like your mind is blank?"
\end{itemize}

\textbf{Scope of This Symptom:}
\begin{itemize}[noitemsep, topsep=0pt]
    \item \textbf{Symptom Reasoning Process:} If the user consistently has trouble concentrating, mark as "Yes".
    \item If the user reports it happening occasionally, mark as "Uncertain".
\end{itemize}

\subsection*{9. Irritability}
\textbf{Symptom Manifestation:} The individual experiences irritability.

\textbf{Corresponding Questions:}
\begin{itemize}[noitemsep, topsep=0pt]
    \item "Do you feel more irritable than usual?"
\end{itemize}

\textbf{Scope of This Symptom:}
\begin{itemize}[noitemsep, topsep=0pt]
    \item \textbf{Symptom Reasoning Process:} If the user expresses irritability consistently, mark as "Yes".
    \item If irritability is occasional, mark as "Uncertain".
\end{itemize}

\subsection*{10. Sleep Disturbance}
\textbf{Symptom Manifestation:} The individual experiences sleep disturbances (difficulty falling asleep, staying asleep, or non-restorative sleep).

\textbf{Corresponding Questions:}
\begin{itemize}[noitemsep, topsep=0pt]
    \item "Do you have trouble sleeping, staying asleep, or feel that your sleep isn't restful?"
\end{itemize}

\textbf{Scope of This Symptom:}
\begin{itemize}[noitemsep, topsep=0pt]
    \item \textbf{Symptom Reasoning Process:} If the user reports any form of sleep disturbance, mark as "Yes".
    \item If the sleep issues are occasional, mark as "Uncertain".
\end{itemize}

\subsection*{11. Impact on Life}
\textbf{Symptom Manifestation:} The anxiety causes significant distress or impairs the individual’s functioning.

\textbf{Corresponding Questions:}
\begin{itemize}[noitemsep, topsep=0pt]
    \item "Has your anxiety caused you significant distress or affected your daily life?"
\end{itemize}

\textbf{Scope of This Symptom:}
\begin{itemize}[noitemsep, topsep=0pt]
    \item \textbf{Symptom Reasoning Process:} If the user reports significant distress or impairment in their life, mark as "Yes".
    \item If the response is unclear or only mildly distressing, mark as "Uncertain".
    \item If no significant impact is reported, mark as "No".
\end{itemize}

\subsection*{\textcolor{red}{Diagnosis Criteria}}
\begin{itemize}[noitemsep, topsep=0pt]
    \item \textbf{Symptom Criteria 1-4:} The user must meet these symptoms ("Yes" for all of these) to diagnose GAD.
    \item \textbf{Symptom Criteria 5-10:} At least three additional symptoms (from 5-10) must be marked "Yes" to confirm the diagnosis.
\end{itemize}

\end{tcolorbox}

\begin{tcolorbox}[title=Social Anxiety Disorder Detection Prompt,enhanced,breakable]

\subsection*{1. Anxiety in Social Situations}
\textbf{Symptom Manifestation:} The individual experiences significant or intense fear or anxiety in social situations where they might be judged by others.

\textbf{Corresponding Questions:}
\begin{itemize}[noitemsep, topsep=0pt]
    \item "Do you feel anxious because of social situations?"
\end{itemize}

\textbf{Scope of This Symptom:}
\begin{itemize}[noitemsep, topsep=0pt]
    \item \textbf{Symptom Reasoning Process:} If the user expresses that anxiety occurs only occasionally, recently, or just for a few days, mark as "No". If the user indicates general, ongoing anxiety in social situations, mark as "Yes".
    \item If the response is ambiguous or no relevant answer is provided, mark as "Uncertain".
\end{itemize}

\subsection*{2. Fear of Social Judgement or Avoidance}
\textbf{Symptom Manifestation:} The individual feels fear or anxiety in almost every social interaction, or avoids social situations due to this fear.

\textbf{Corresponding Questions:}
\begin{itemize}[noitemsep, topsep=0pt]
    \item "Does the fear almost always affect you when you're in social situations?"
    \item "Do you avoid social activities because of fear?"
\end{itemize}

\textbf{Scope of This Symptom:}
\begin{itemize}[noitemsep, topsep=0pt]
    \item \textbf{Symptom Reasoning Process:} If the individual experiences anxiety in almost all social situations or avoids them altogether, mark as "Yes".
    \item Otherwise, mark as "No".
\end{itemize}

\subsection*{3. Excessive Anxiety}
\textbf{Symptom Manifestation:} The fear or anxiety is disproportionate to the actual threat or situation.

\textbf{Corresponding Questions:}
\begin{itemize}[noitemsep, topsep=0pt]
    \item "Do you feel that your anxiety or fear in social situations is more than what's actually happening?"
\end{itemize}

\textbf{Scope of This Symptom:}
\begin{itemize}[noitemsep, topsep=0pt]
    \item \textbf{Symptom Reasoning Process:} If the user feels that their anxiety is excessive compared to the actual threat, mark as "Yes".
    \item Otherwise, mark as "No".
    \item If the response is unclear, mark as "Uncertain".
\end{itemize}

\subsection*{4. Avoidance Behavior}
\textbf{Symptom Manifestation:} The individual avoids social interactions or seeks accompaniment due to fear or anxiety.

\textbf{Corresponding Questions:}
\begin{itemize}[noitemsep, topsep=0pt]
    \item "Do you actively avoid social situations or prefer to go with someone?"
\end{itemize}

\textbf{Scope of This Symptom:}
\begin{itemize}[noitemsep, topsep=0pt]
    \item \textbf{Symptom Reasoning Process:} If the user avoids social situations or requires accompaniment frequently, mark as "Yes".
    \item Otherwise, mark as "No".
    \item If the response is unclear, mark as "Uncertain".
\end{itemize}

\subsection*{5. Impact on Life}
\textbf{Symptom Manifestation:} The symptoms cause significant distress or interfere with daily functioning (e.g., work, social life, school, relationships).

\textbf{Corresponding Questions:}
\begin{itemize}[noitemsep, topsep=0pt]
    \item "Have these symptoms caused you significant distress or affected your daily life?"
    \item "Do these symptoms impact your ability to work, study, or have social relationships?"
\end{itemize}

\textbf{Scope of This Symptom:}
\begin{itemize}[noitemsep, topsep=0pt]
    \item \textbf{Symptom Reasoning Process:} If the user reports significant distress or impairment in their daily life due to these symptoms, mark as "Yes".
    \item If the user indicates no significant impact, mark as "No".
    \item If the response is unclear, mark as "Uncertain".
\end{itemize}

\subsection*{\textcolor{red}{Diagnosis Criteria}}
\begin{itemize}[noitemsep, topsep=0pt]
    \item \textbf{Symptoms 1-4:} If at least one symptom is marked "No", the diagnosis is "No".
    \item \textbf{Symptom 5:} If "Yes" is indicated for significant distress or impairment, mark as "Yes".
    \item If the individual meets all the criteria and shows significant distress or impairment, mark as "Yes". Otherwise, mark as "No".
\end{itemize}

\end{tcolorbox}

\begin{tcolorbox}[title=Suicide Detection Prompt,enhanced,breakable]

\subsection*{1. Suicide Ideation}
\textbf{Symptom Manifestation:} The individual has thoughts of ending their own life, including mild or intense suicidal thoughts, thoughts about death, or expressions of hopelessness.

\textbf{Corresponding Questions:}
\begin{itemize}[noitemsep, topsep=0pt]
    \item "Have you had any thoughts about ending your life or about death in general?"
    \item "Do you ever feel hopeless or like life isn't worth living?"
\end{itemize}

\textbf{Scope of This Symptom:}
\begin{itemize}[noitemsep, topsep=0pt]
    \item \textbf{Symptom Reasoning Process:} If the individual reports thoughts about death or suicide, mark as "Yes".
    \item If there is no mention of suicidal ideation or thoughts of death, mark as "No".
    \item If the response is unclear or related to mild, transient thoughts, mark as "Uncertain".
\end{itemize}

\subsection*{2. Suicide Plan}
\textbf{Symptom Manifestation:} The individual has a detailed plan to carry out suicide, including specific methods, times, locations, and preparations (e.g., buying pills, writing a will).

\textbf{Corresponding Questions:}
\begin{itemize}[noitemsep, topsep=0pt]
    \item "Have you made any specific plans for how you would end your life?"
    \item "Do you have a method or any preparations in place for suicide?"
\end{itemize}

\textbf{Scope of This Symptom:}
\begin{itemize}[noitemsep, topsep=0pt]
    \item \textbf{Symptom Reasoning Process:} If the individual describes having a specific plan, such as a method or location, mark as "Yes".
    \item If there is no mention of a detailed plan or preparation, mark as "No".
    \item If the response is vague or there is uncertainty, mark as "Uncertain".
\end{itemize}

\subsection*{3. Suicide Attempt}
\textbf{Symptom Manifestation:} The individual has taken action toward committing suicide, including behaviors such as overdosing, hanging, jumping, etc., with a high risk of lethality.

\textbf{Corresponding Questions:}
\begin{itemize}[noitemsep, topsep=0pt]
    \item "Have you ever tried to take your life or attempted suicide?"
    \item "Have you ever engaged in any behavior that could harm yourself severely?"
\end{itemize}

\textbf{Scope of This Symptom:}
\begin{itemize}[noitemsep, topsep=0pt]
    \item \textbf{Symptom Reasoning Process:} If the individual reports taking significant steps toward suicide, such as attempting overdose, hanging, or other lethal actions, mark as "Yes".
    \item If the individual denies any past suicide attempt or dangerous behavior, mark as "No".
    \item If the response is unclear or ambiguous, mark as "Uncertain".
\end{itemize}

\subsection*{\textcolor{red}{Diagnosis Criteria}}
\begin{itemize}[noitemsep, topsep=0pt]
    \item If any of the three symptoms (suicide ideation, plan, or attempt) is marked as "Yes" and the severity is mild, moderate, or severe, mark the diagnosis as "Yes".
    \item If there is no mention of suicide-related behaviors or thoughts, or if the content is irrelevant, mark as "No".
\end{itemize}

\end{tcolorbox}

\twocolumn